%% file: main.tex
\documentclass{article} 
\usepackage{iclr2017_conference,times}
\usepackage{hyperref}
\usepackage{url}
\usepackage{color}
\usepackage{comment}
\usepackage{soul}
\usepackage{soul}
\usepackage{ragged2e}
\usepackage{placeins}
\usepackage{graphicx}

\usepackage{amsfonts}
\title{Dialogue Learning With Human-in-the-Loop}
\author{Jiwei Li, Alexander H. Miller, Sumit Chopra, Marc'Aurelio Ranzato, Jason Weston \\
Facebook AI Research, \\
New York, USA \\
\texttt{\{jiwel,ahm,spchopra,ranzato,jase\}@fb.com} \\
}

\newcommand{\MINUS}{}

\begin{document}

\maketitle

\begin{abstract}
An important aspect of developing conversational agents is 
to give a bot the 
ability to improve 
through communicating with humans and to learn from the mistakes that it makes.  
Most research has focused on learning
from fixed training sets of labeled data rather than interacting with a dialogue partner
 in an online fashion.
In this paper we explore this direction in a reinforcement learning setting where the bot
improves its question-answering ability
from feedback a teacher
gives following its generated responses.
We build a simulator that tests various aspects of such learning in a synthetic environment,
and introduce models that work in this regime. 
Finally, real experiments with Mechanical Turk validate the approach.
\end{abstract}

\section{Introduction}
\input{intro}

\section{Related Work} \label{sec:related}
\input{related}
\section{Dataset and Tasks}
\input{data}

\section{Methods}

\subsection{Model Architecture}
\input{model}

\subsection{Reinforcement Learning} \label{sec:rl} 
\input{methods}

\section{Experiments}
\input{exp}

\section{Conclusion}
\input{conc}

\bibliography{iclr2017_conference}
\bibliographystyle{iclr2017_conference}

\newpage
\appendix

\section{Further Simulator Task Details} \label{sec:extra_data}
\input{extra_data}

\section{Instructions given to Turkers} \label{sec:mturk}
\input{mturk}

\newpage
\section{Additional Experiments}
\input{extra_exp}

\end{document}

%% file: intro.tex
A good conversational agent (which we sometimes refer to as a learner or bot\footnote{In this paper, we refer to a learner (either a human or a bot/dialogue agent which is a machine learning algorithm) as the student, and their more knowledgeable dialogue partner as the teacher.})
should have the ability to learn from the online feedback from a teacher: adapting its model when making mistakes and reinforcing the model when the teacher's feedback is positive.
This is particularly important in the situation where the bot is initially trained in a supervised way on a fixed synthetic, domain-specific or pre-built dataset before release, but will be exposed to
a different environment after release
 (e.g.,
 more diverse natural language utterance usage when talking with real humans, different distributions, special cases, etc.).  Most recent research
has focused on training a bot
from fixed training sets of labeled data but seldom on how the bot can
improve through online interaction with humans.
Human (rather than machine) language learning happens during communication \citep{bassiri2011interactional,werts1995instructive},
and not from labeled datasets, hence making this an important subject to study.

In this work, we explore this direction by
 training a bot through interaction with teachers in an online fashion.
 The task is formalized under the general framework of reinforcement learning
via the teacher's (dialogue partner's) feedback to the dialogue actions from the bot.
 The dialogue takes place in the context of question-answering tasks and the bot has to, given either a short story or a set of facts, answer a set of questions from the teacher.
We consider two types of feedback: explicit numerical rewards as in conventional
reinforcement learning, and textual feedback which is more natural in human dialogue, following
\citep{weston2016dialog}.
We consider two online training scenarios:
(i) where the task is built with a dialogue simulator allowing for easy analysis and repeatability
of experiments; and (ii) where the teachers are real humans using Amazon Mechanical Turk.

    We explore  important issues involved in online learning
   such as  how a bot can be most efficiently trained using a minimal amount of teacher's feedback,
 how a bot can harness different types of feedback signal,
how to avoid pitfalls such as instability during online learing with different types of feedback via
 data balancing and exploration,
and how to make learning with real humans feasible via data batching.
Our findings indicate that
it is feasible to build a pipeline
that starts from a model trained with fixed data and then learns from interactions with humans
to improve itself. 

%% file: related.tex
Reinforcement learning has been widely applied to dialogue, especially in slot filling
 to solve
domain-specific tasks \citep{walker2000application,schatzmann2006survey,singh2000empirical,singh2002optimizing}.
Efforts include Markov Decision Processes
 (MDPs) \citep{levin1997learning,levin2000stochastic,walker2003trainable,pieraccini2009we},
POMDP models \citep{young2010hidden,young2013pomdp,gavsic2013pomdp,gavsic2014incremental} and
policy learning \citep{su2016continuously}.
Such a line of research focuses mainly on frames with slots to fill, where the
bot will use reinforcement learning to model a state transition pattern,
generating dialogue utterances to prompt the appropriate user responses to put in
the desired slots.
This goal is different from ours, where we study end-to-end learning systems and
also consider non-reward based setups via textual feedback.

Our work is related to the line of research that focuses on supervised learning
for question answering (QA) from dialogues
\citep{dodge2015evaluating,weston2016dialog}, either given a database of knowledge \citep{bordes2015large,miller2016key}
 or short texts \citep{weston2015towards,hermann2015teaching,rajpurkar2016squad}. 
In our work, the discourse includes the statements made in the past, the question and answer, 
and crucially the response from the teacher. 
The latter is what makes the setting different from the standard QA setting, i.e. 
we use methods that leverage this response also, not just answering questions.
Further, QA works only consider fixed datasets with gold annotations, 
i.e. they  do not consider a reinforcement learning setting.

Our work is closely related to a recent work from \cite{weston2016dialog}
that learns  through conducting conversations where supervision is given 
naturally in the response during the conversation.
 That work introduced the use 
of forward prediction that learns by predicting the teacher's feedback,
in addition to using reward-based learning of correct answers.
However,  two important issues were not addressed:
(i) it did not use a reinforcement learning setting, but instead used
pre-built datasets with fixed policies given in advance;
and (ii) experiments used only simulated and no real language data.
Hence, models that can learn policies from real online communication were not investigated.
To make the differences with our work clear, we will now detail these points further.

%
%
The experiments in \citep{weston2016dialog} involve constructing pre-built fixed datasets, rather
than training the learner within a simulator, as in our work.
Pre-built datasets can only be made by fixing a prior in advance.
They achieve this by choosing an omniscient (but deliberately imperfect) labeler that gets $\pi_{acc}$
examples always correct (the paper looked at values 50\%, 10\% and 1\%).
Again, this was not learned, and was fixed to generate the datasets.
Note that the paper refers to these answers as coming from ``the learner'' (which should be the model), but since the policy is fixed it actually does not depend on the model.
 In a realistic setting one does not have access to an omniscient labeler, one has to
 learn a policy completely from scratch, online, starting with a random policy, so their setting was not
practically viable. In our work, 
when policy training is viewed as batch learning over iterations of the dataset, updating
the policy on each iteration, \citep{weston2016dialog} can be viewed as training only one iteration, whereas
 we perform multiple iterations. This is explained further in
 Sections \ref{sec:rl} and \ref{sec:online_exp}.
We show in our experiments that performance improves over the iterations, i.e. it is better than the
first iteration. We show that such online learning works for both reward-based numerical feedback
 and for forward prediction methods using textual feedback (under certain conditions which are detailed).
This is a key contribution of our work.

Finally, \citep{weston2016dialog} only conducted experiments on synthetic or templated language, and not real
language, especially the feedback from the teacher was scripted. While we believe that synthetic datasets
are very important for developing understanding (hence we develop a simulator and conduct experiments also with synthetic data), for a new method to gain traction it must be shown to work on real data.
We hence employ Mechanical Turk to collect real language data for the questions and importantly for the 
teacher feedback and construct experiments in this real setting.

%% file: data.tex
We begin by describing the data setup we use.
In our first set of experiments we build a simulator as a testbed for learning algorithms.
In our second set of experiments we use Mechanical Turk to provide real human teachers giving feedback.


\subsection{Simulator}

The simulator adapts two existing fixed datasets to our online setting.
Following \cite{weston2016dialog}, we use
(i) the single supporting fact problem from the bAbI datasets \citep{weston2015towards}
which consists of 1000 short stories from a simulated world interspersed with questions; and (ii)
the WikiMovies dataset \citep{weston2015towards} which consists
 of roughly 100k (templated) questions
over 75k entities based on questions with answers in the open movie database (OMDb).
Each dialogue takes place between a teacher, scripted by the simulation, and a bot.
The communication protocol is as follows: (1)
the teacher first asks a question from the fixed set of questions existing in the dataset, (2)
the bot answers the question, and finally (3) the teacher gives feedback on the bot's answer.

 We follow the paradigm defined in \citep{weston2016dialog} where the teacher's feedback
 takes the form of either textual feedback, a numerical reward, or both, depending on the task.
For each dataset, there are ten tasks,
which are further described in Sec.~\ref{sec:extra_data} and illustrated in Figure~\ref{Tasks} of the appendix.
We also refer the readers to \citep{weston2016dialog} for more detailed descriptions and
the motivation behind these tasks.
In the main text of this paper we
only consider 
 Task 6 (``partial feedback''): 
the teacher replies with positive textual feedback (6 possible templates) when the bot answers correctly, and positive reward is given only 50\% of the time.
When the bot is wrong, the teacher gives textual feedback containing the answer.
Descriptions and experiments on the other tasks are detailed in the appendix.
Example dialogues are given in Figure \ref{fig:simulator-examples}.

The difference between our simulation and the original fixed tasks of~\cite{weston2016dialog}
is that models are trained on-the-fly. 
After receiving feedback and/or rewards, we update the model (policy) and then deploy it to collect
teacher's feedback in the next episode or batch.
This means the model's policy affects the data which is used to train it,
which was not the case in the previous work.

\definecolor{dred}{rgb}{0.7,0.0,0.0}
\newcommand{\PLUS}{{\textcolor{blue}{(+)}}}
\newcommand{\SPACE}{~~~~~~~~~~~~~~~~~~~~~~}
\begin{figure*}[h]
\begin{small}
\caption{{\bf Simulator sample dialogues for the bAbI task (left) and WikiMovies (right).}
We consider 10 different tasks following \cite{weston2016dialog} but here describe only Task 6;
other tasks are detailed in the appendix.
The teacher's dialogue is in black and the bot is in red.
$\PLUS$ indicates receiving positive reward, given only 50\% of the time even when correct.
\label{fig:simulator-examples}
}
\begin{tabular}{|l|c|l|}
\cline{1-1}\cline{3-3}
&& \\[-2ex]
{\bf bAbI Task 6: Partial Rewards} &&
{\bf WikiMovies Task 6: Partial Rewards}  \\
&& \\[-2ex]
Mary went to the hallway.         &&
What films are about Hawaii? \SPACE \textcolor{dred}{50 First Dates}\\
John moved to the bathroom.    &&
Correct! \\  
Mary travelled to the kitchen.     &&
Who acted in Licence to Kill? \SPACE \textcolor{dred}{Billy Madison}\\
Where is Mary? \SPACE~ \textcolor{dred}{kitchen} &&
No, the answer is Timothy Dalton.\\
Yes, that's right! &&
What genre is Saratoga Trunk in?  \SPACE    \textcolor{dred}{Drama}\\
Where is John? \SPACE~~ \textcolor{dred}{bathroom}  &&
Yes! \PLUS \\
Yes, that's correct!  \PLUS   &&  ~~~~~~\dots  \\
\cline{1-1}\cline{3-3}
\end{tabular}
\end{small}
\end{figure*}
\begin{figure*}[h]
\center
\begin{small}
\caption{{\bf Human Dialogue from Mechanical Turk (based on WikiMovies)}
The human teacher's dialogue is in black and the bot is in red.
We show examples where the bot answers correctly (left) and incorrectly (right).
Real humans provide more variability of language in both questions and textual feedback
than in the simulator setup (cf. Figure \ref{fig:simulator-examples}).
 \label{fig:mturk_data}}\vspace{2mm}
\begin{tabular}{|l|}
\hline
{\bf Sample dialogues with correct answers from the bot:}\\
Who wrote the Linguini Incident ?     \SPACE~~~~~~~~~~~~~~~~~~~ \textcolor{dred}{richard shepard}\\
Richard Shepard is one of the right answers here.\\
What year did The World Before Her premiere? \SPACE \textcolor{dred}{2012}\\
Yep! That's when it came out.  \\
Which are the movie genres of Mystery of the 13th Guest?  ~~~ \textcolor{dred}{crime}\\
\vspace{1mm}
Right, it can also be categorized as a mystery.\\
{\bf Sample dialogues with incorrect answers from the bot:}\\
What are some movies about a supermarket ?    \SPACE~~~ \textcolor{dred}{supermarket} \\
There were many options and this one was not among them.  \\
Which are the genres of the film Juwanna Mann ? ~~~~~~~~~~~~~~~~~~~ \textcolor{dred}{kevin pollak} \\
That is incorrect. Remember the question asked for a genre not name. \\
Who wrote the story of movie Coraline ?    \SPACE~~~~~~~~~~~~\textcolor{dred}{fantasy} \\
That's a movie genre and not the name of the writer. A better answer would of been Henry Selick\\
 or Neil Gaiman.\\
\hline
\end{tabular}
\end{small}
\vspace*{-3ex}
\end{figure*}

\subsection{Mechanical Turk Experiments} \label{sec:data-mturk}

Finally, we extended WikiMovies using Mechanical
Turk so that real human teachers are  giving feedback rather than using a simulation.
As both the questions and feedback are templated in the simulation, they are now both
replaced with natural human utterances. Rather than having a set of simulated tasks, we have
only one task, and we gave instructions to the teachers that they could give feedback
as they see fit. The exact instructions given to the Turkers is given in Appendix \ref{sec:mturk}.
In general, each independent response  contains feedback like
(i) positive or negative sentences; or (ii) a phrase containing the answer
or (iii) a hint,  which are similar to setups defined in the simulator. However,
 some human responses cannot be so easily categorized,
and the lexical variability is much larger in human responses.
Some examples of the collected data are given in Figure~\ref{fig:mturk_data}.

%% file: model.tex
In our experiments, we used variants of the
End-to-End Memory Network  (MemN2N) model \citep{sukhbaatar2015end} 
as our underlying architecture for learning from dialogue. 

The input to MemN2N is the last utterance of the dialogue history $x$ as well as a set of memories (context)
 $C$=$c_1$, $c_2$, ..., $c_N$.
The memory $C$ encodes  both short-term memory, e.g., dialogue histories between the bot and the teacher,
 and long-term memories, e.g., the knowledge base facts that the bot has access to.
Given the input $x$ and $C$, the goal is to produce an output/label $a$.

In the first step, the query $x$ is transformed to a vector representation $u_0$  by summing up its constituent word embeddings: $u_0=Ax$. The input $x$ is a bag-of-words vector and $A$ is the $d\times V$ word embedding matrix where $d$ denotes the emebbding dimension and $V$\
 denotes the vocabulary size. Each memory $c_i$ is similarly transformed to a vector $m_{i}$.
The model will read information from the memory by comparing
 input representation $u_0$ with memory vectors $m_{i}$ using softmax weights:
\begin{equation}
o_1=\sum_{i}p_i^1 m_{i} ~~~~~~~~~~p_i^1=\texttt{softmax}(u_0^T m_i)
\end{equation}
This process selects memories relevant to the last utterance $x$,
i.e., the memories with large values of $p_i^1$.
The returned memory vector $o_1$ is the weighted sum of memory vectors.
This process can be repeated to query the memory N times (so called ``hops'')
by adding $o_n$ to the original input, $u_1=o_1+u_0$, or to the previous state, $u_n=o_n+u_{n-1}$,
and then using $u_n$ to query the memories again.

In the end, $u_N$ is input to a softmax function for the final prediction:
\begin{equation}\label{eq:a}
a=\texttt{softmax} (u_N^T y_1,u_N^T y_2,...,u_N^T y_L)
\end{equation}
where $y_1, \dots, y_L$ denote the set of candidate answers.
If the answer is a word, $y_i$ is the corresponding word embedding.
If the answer is a sentence, $y_i$ is the embedding for the sentence achieved in
the same way that we obtain embeddings for query $x$ and memory $C$.

The standard way MemN2N is trained is via a cross entropy criterion on known input-output pairs,
which we refer to as supervised or imitation learning.
As our work is in a reinforcement learning setup where our model must make predictions to learn,
this procedure will not work, so we instead consider reinforcement learning algorithms which we describe next.

%% file: methods.tex
In this section, we present the algorithms we used to train MemN2N in an online fashion.
Our learning setup can be cast as a particular form of Reinforcement Learning. 
The policy is implemented by the MemN2N model. 
The state is the dialogue history. 
The action space
corresponds to the set of answers the MemN2N has to choose from to answer the teacher's
 question. In our setting, the policy chooses only one action for each episode.
The reward is either $1$ (a reward from the teacher when the bot answers correctly) 
 or $0$ otherwise.
Note that in our experiments, a reward equal to $0$ might mean that the answer is incorrect or
that the positive reward is simply missing.
The overall setup is closest to standard contextual bandits, except that the reward is binary.

When working with real human dialogues, e.g. collecting data via Mechanical Turk,
it is easier to set up a task whereby a bot is deployed to respond to a large batch of utterances,
as opposed to a single one.
The latter would be more difficult to manage and scale up since it would require some
form of synchronization between the model replicas interacting with each human.

This is comparable to the real world situation where a teacher can either ask a student a single question and give feedback right away,
or set up a test that contains many questions and grade all of them at once. Only after the learner completes all questions, it can hear
feedback from the teacher.

We use {\it batch size} to refer to how many dialogue
episodes the current model is used to collect feedback before updating its parameters.
In the Reinforcement Learning literature, batch size is related to {\em off-policy}
learning since the MemN2N policy is trained
using episodes collected with a stale version of the model. Our experiments show
that our model and base algorithms are very robust to the choice
of batch size, alleviating the need for correction terms in the learning algorithm~\citep{bottou-13}.

We consider two strategies: (i) online batch size, whereby the target policy is
updated after doing a single pass over each batch (a batch size of 1 reverts to
the usual on-policy online learning); and (ii) dataset-sized batch, whereby
training is continued to convergence on the batch which is the size of the dataset,
and then the target policy is updated with the new model, and a new batch is drawn and the procedure iterates.
These strategies can be applied to all the methods we use, described below.
%

Next, we discuss the learning algorithms we considered in this work.

\subsubsection{Reward-Based Imitation (RBI)}
The simplest algorithm we first consider is the one employed in \cite{weston2016dialog}.
RBI relies on positive rewards provided by the teacher.
It is trained to imitate the correct behavior of the learner, i.e.,
learning to predict the correct answers (with reward 1) at training time and disregarding the other ones.
This is implemented by using a {MemN2N} that maps a dialogue input to a prediction, i.e. 
using the cross entropy criterion on the positively rewarded subset of the data.

In order to make this work in the online setting which requires exploration to find the correct answer,
we employ an $\epsilon$-greedy strategy:
the learner makes a prediction using its own model (the answer assigned the highest probability)
 with probability $1-\epsilon$, otherwise it picks a random answer with probability $\epsilon$.
The teacher will then give a reward of $+1$ if the answer is correct, otherwise $0$.
The bot will then learn to imitate the correct answers:
predicting the correct answers while ignoring the incorrect ones.

\subsubsection{REINFORCE}
The second algorithm we use is the REINFORCE algorithm \citep{williams1992simple},
which maximizes the expected cumulative reward of the episode, in our case the expected reward provided by the teacher.
The expectation is approximated by sampling an answer from the model distribution.
Let $a$ denote the answer that the learner gives,
$p(a)$ denote the probability that current model assigns to $a$,
$r$ denote the teacher's reward, and $J(\theta)$ denote the expectation of the reward. We have:
\begin{equation}
\nabla J(\theta)\approx\nabla\log p(a) [r-b]
\end{equation}
where $b$ is the baseline value, which is estimated using a linear regression model
that takes as input the output of the memory network after the last hop,
and outputs a scalar $b$ denoting the estimation of the future reward.
The baseline model is trained by minimizing the mean squared loss between the estimated reward $b$ and actual reward $r$, $||r-b||^2$.
We refer the readers to \citep{ranzato2015sequence,zaremba2015reinforcement} for more details.
The baseline estimator model is independent from the policy model, and its error is not backpropagated through the policy model.

The major difference between RBI and REINFORCE is that (i) the learner only tries to imitate correct behavior in RBI while in REINFORCE it also leverages the incorrect behavior,
and (ii) the learner explores using an $\epsilon$-greedy strategy in RBI while in REINFORCE it uses the distribution over actions produced by the model itself.

\subsubsection{Forward Prediction (FP)}
FP \citep{weston2016dialog} handles the situation where a numerical reward for a bot's answer is not available, meaning that there are no +1 or 0 labels available after a student's utterance.
Instead, the model assumes
 the teacher gives  textual feedback $t$ to the bot's answer, taking the form of a dialogue utterance,
and the model tries to predict this instead.
Suppose that $x$ denotes the teacher's question and $C$=$c_1$, $c_2$, ..., $c_N$ denotes the dialogue history as before.
In {\it FP}, the model first maps the teacher's initial question $x$ and dialogue history $C$ to
a vector representation $u$ using a memory network with multiple hops.
Then the model will perform another hop of attention over all possible student's answers  in $\mathbb{A}$, 
with an additional part that incorporates the information
of which candidate (i.e., $a$) was actually selected in the dialogue:
\begin{equation}
p_{\hat{a}}=\texttt{softmax}(u^T y_{\hat{a}})~~~~~o=\sum_{\hat{a}\in \mathbb{A}} p_{\hat{a}} (y_{\hat{a}}+\beta\cdot {\bf 1}[\hat{a}=a] )
\end{equation}
where $y_{\hat{a}}$ denotes the vector representation for the student's  answer candidate $\hat{a}$.
$\beta$ is a (learned) d-dimensional vector to signify the actual action $a$ that the student chooses.
$o$ is then combined with $u$ to predict the teacher's feedback $t$ using a softmax:
\begin{equation}
u_1=o+u ~~~~
t=\texttt{softmax} (u_1^T x_{r_1}, u_1^T x_{r_2}, ..., u_1^T x_{r_N})
\end{equation}
where $x_{r_{i}}$ denotes the embedding for the $i^{th}$ response.
In the online setting, the teacher will give textual feedback,
and the learner needs to update its model using the feedback.
It was shown in \cite{weston2016dialog} that in an off-line setting this procedure can work either on its own, or in conjunction with a method that uses numerical rewards as well for improved performance.
In the online setting,  we consider two simple extensions:
\begin{itemize}
\item $\epsilon$-greedy exploration:  with probability $\epsilon$
the student will give a random answer, and with probability $1-\epsilon$ it
will give the answer that its model assigns the largest probability. This method enables the model to explore the space of actions and to potentially discover correct answers.
\item data balancing: cluster the set of teacher responses $t$ and then balance training across the clusters equally.\footnote{In the simulated data, because the responses are templates, this can be implemented by first randomly sampling the response, and then randomly sampling a story with that response; we keep the history of all stories seen from which we sample. For real data slightly more sophisticated clustering should be used.}
This is a type of experience replay \citep{mnih2013playing} but sampling with an evened distribution.
Balancing stops part of the distribution dominating the learning.
For example, if the  model is not exposed to sufficient positive and negative feedback,
and one class overly dominates, the learning process degenerates
to a model that always predicts the same output regardless of its input.

\end{itemize}


%% file: exp.tex
Experiments are first conducted using our simulator, and then
using Amazon Mechanical Turk with real human subjects taking the role of the teacher\footnote{
Code and data are available at
\tiny{\url{https://github.com/facebook/MemNN/tree/master/HITL}}.
}.

\begin{figure*}[!ht]
\includegraphics[width=2.5in]{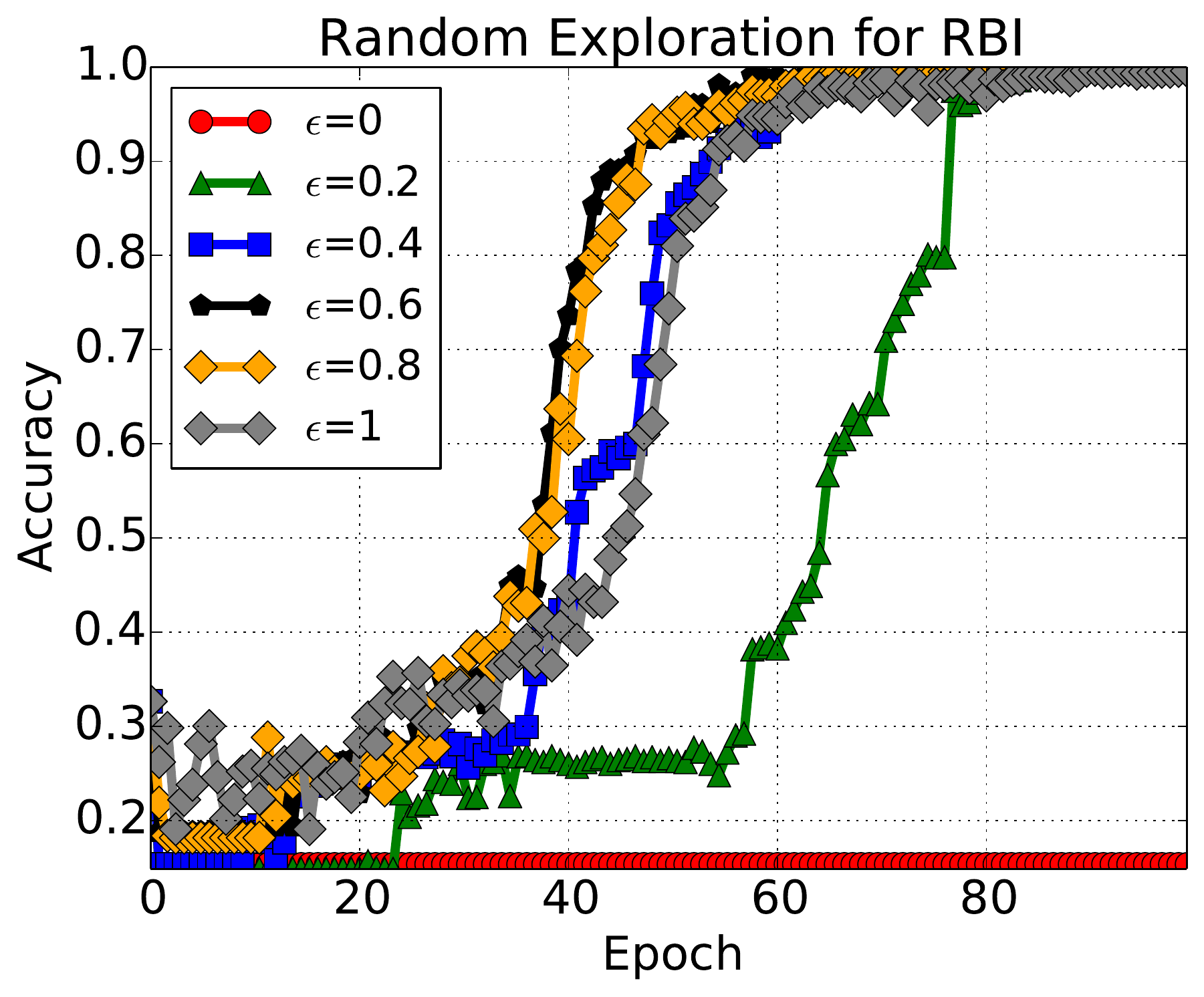}
\includegraphics[width=2.5in]{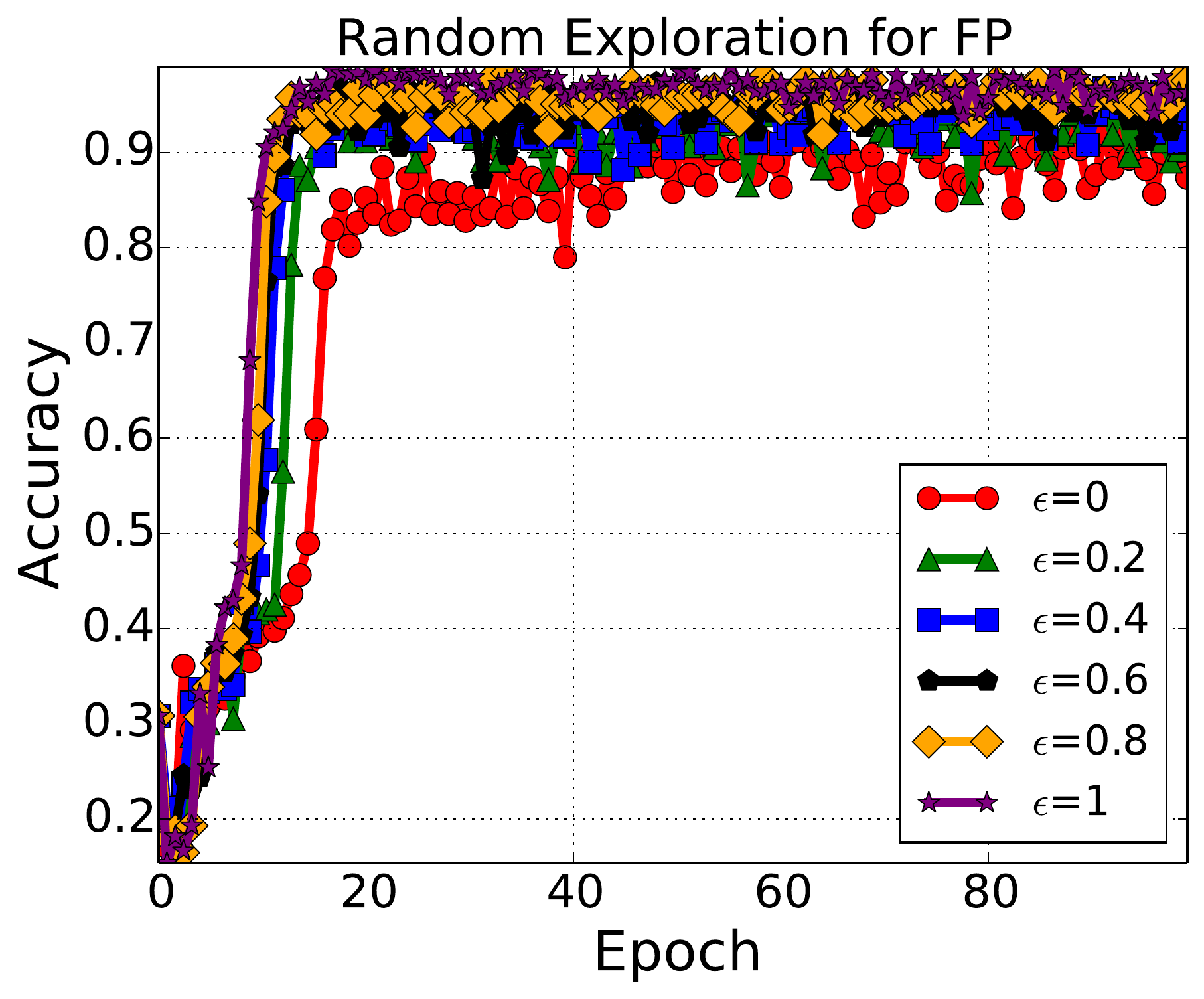}\\
\includegraphics[width=2.5in]{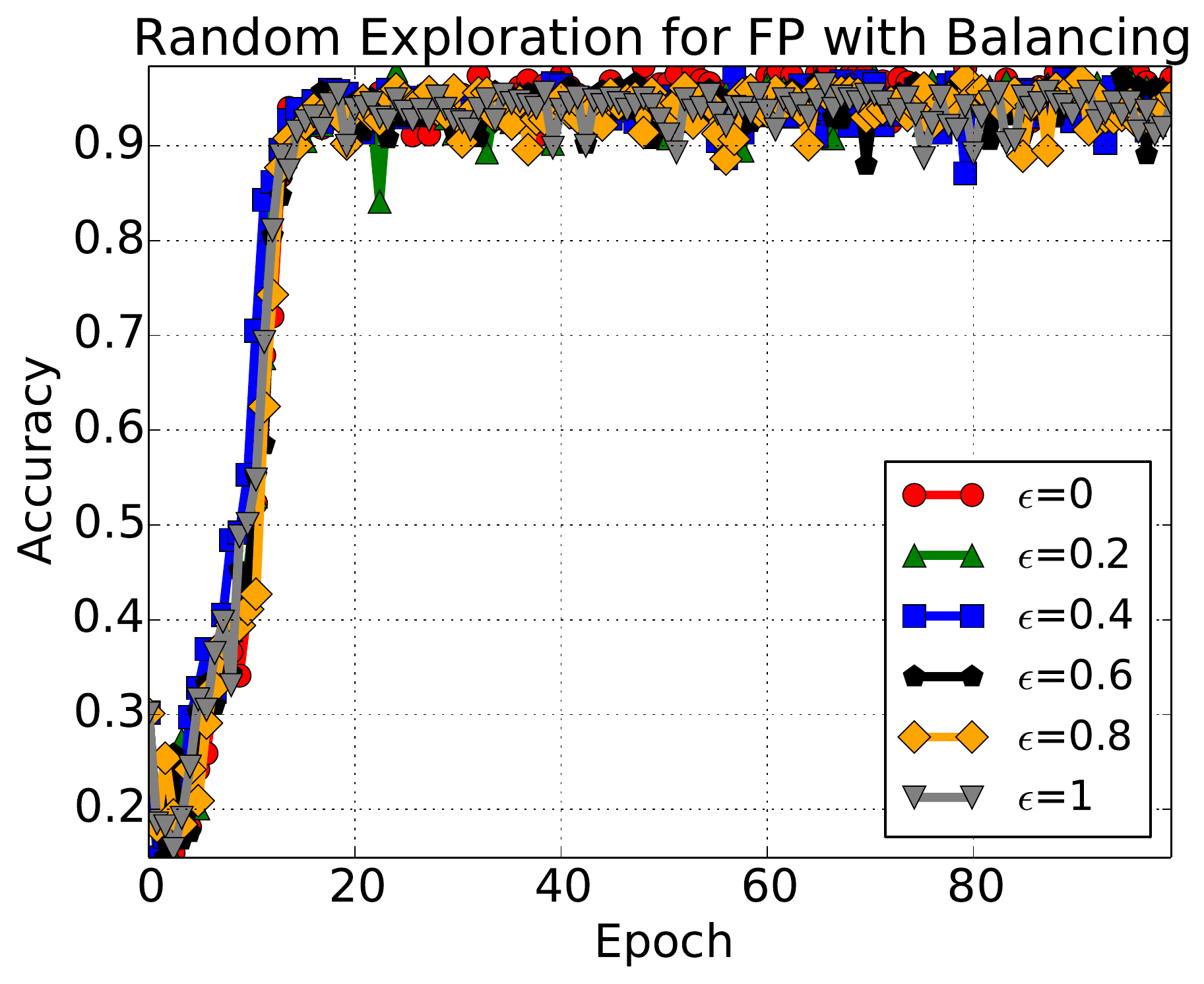}
\includegraphics[width=2.5in]{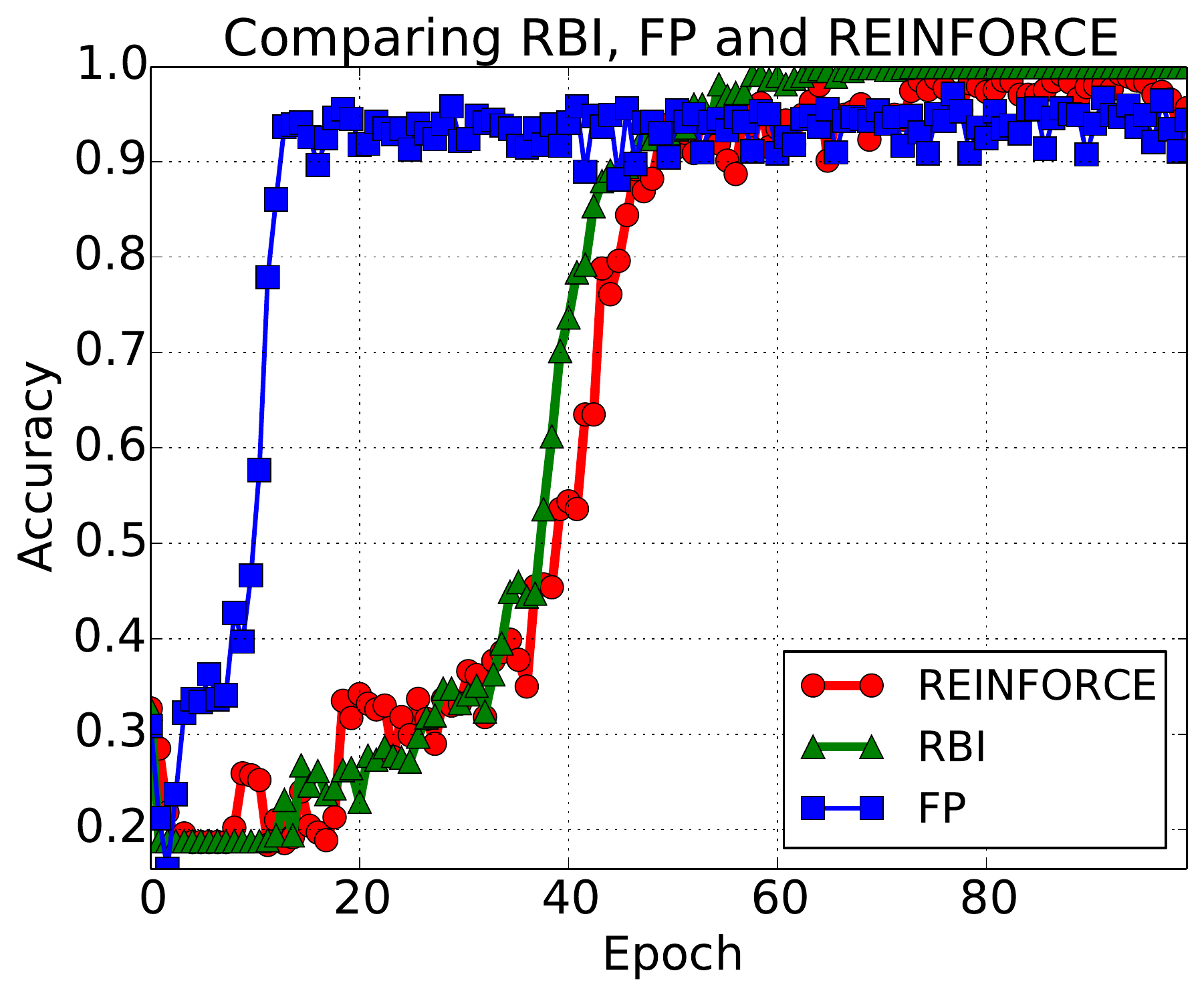}\\
\includegraphics[width=2.5in]{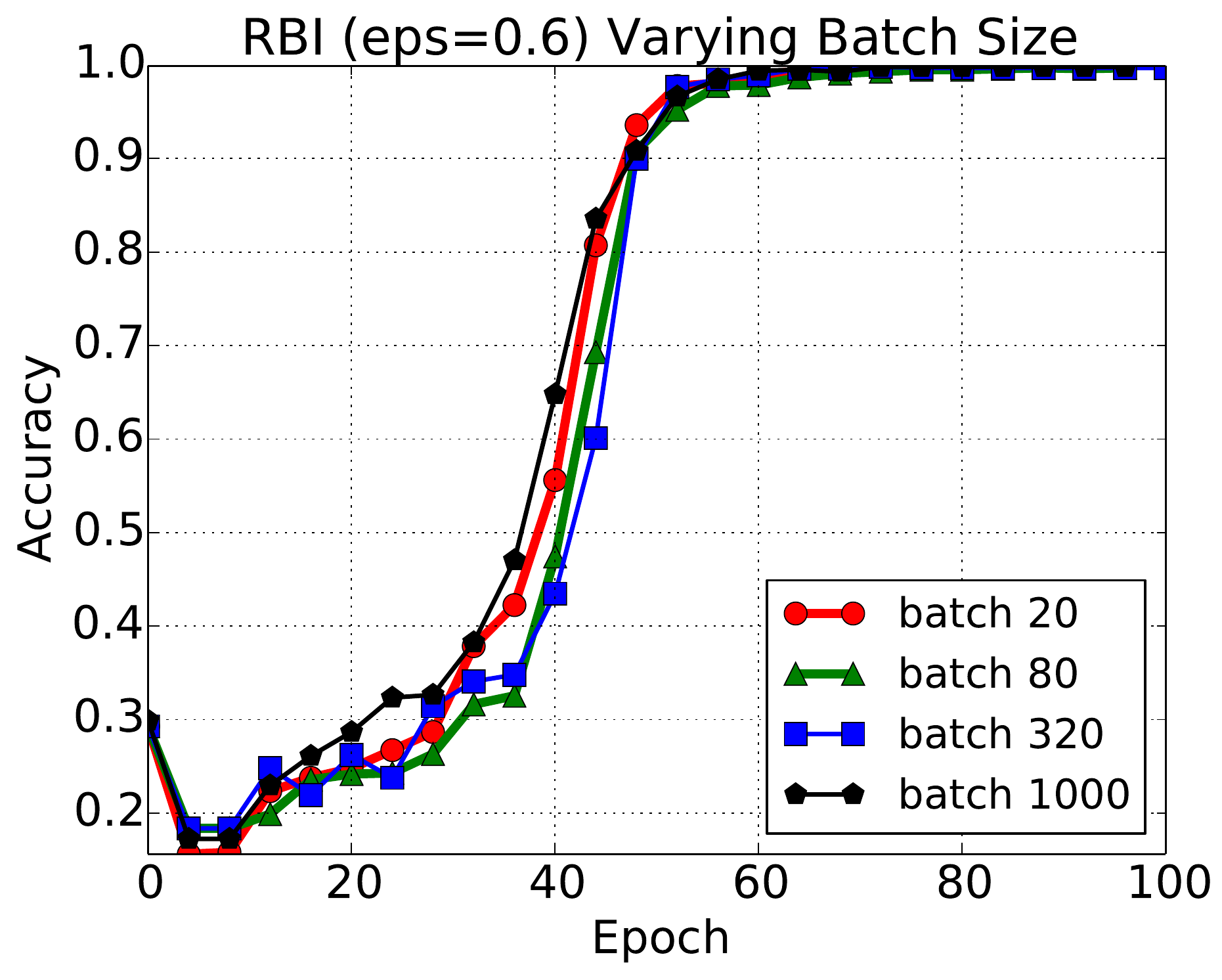}
\includegraphics[width=2.5in]{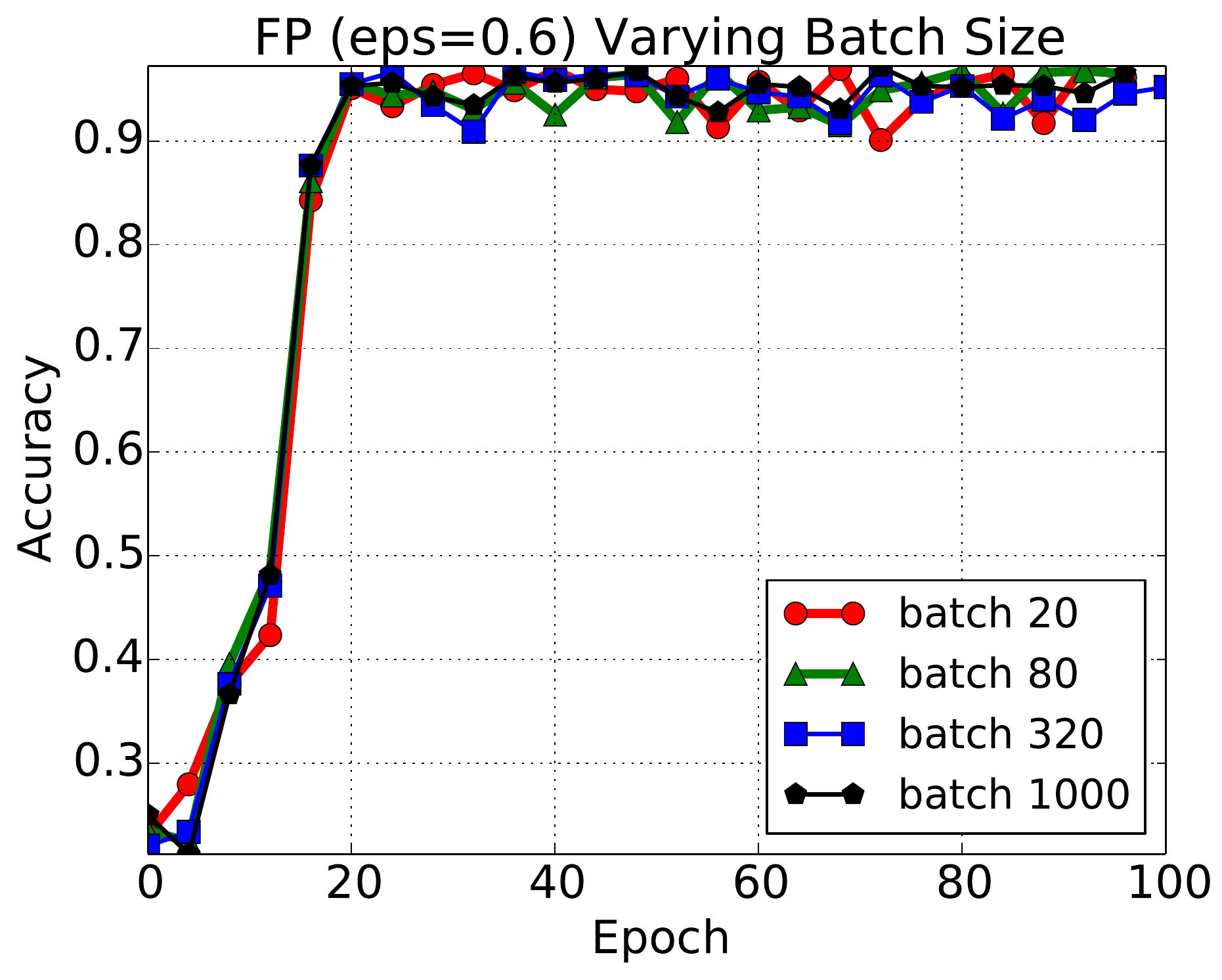}\\
\caption{{\bf  Training epoch vs. test accuracy for bAbI (Task 6) varying exploration $\epsilon$ and batch size.}
Random exploration is important for both reward-based (RBI) and forward prediction (FP).
Performance is largely independent of batch size, and
RBI performs similarly to REINFORCE. 
Note that supervised, rather than reinforcement learning, with gold standard labels
achieves 100\% accuracy on this task.
\label{fig:online-babi-task6}
}
\end{figure*}

\subsection{Simulator}

\paragraph{Online Experiments} \label{sec:online_exp}

\begin{figure*}[!t]
\center
\includegraphics[width=2.35in]{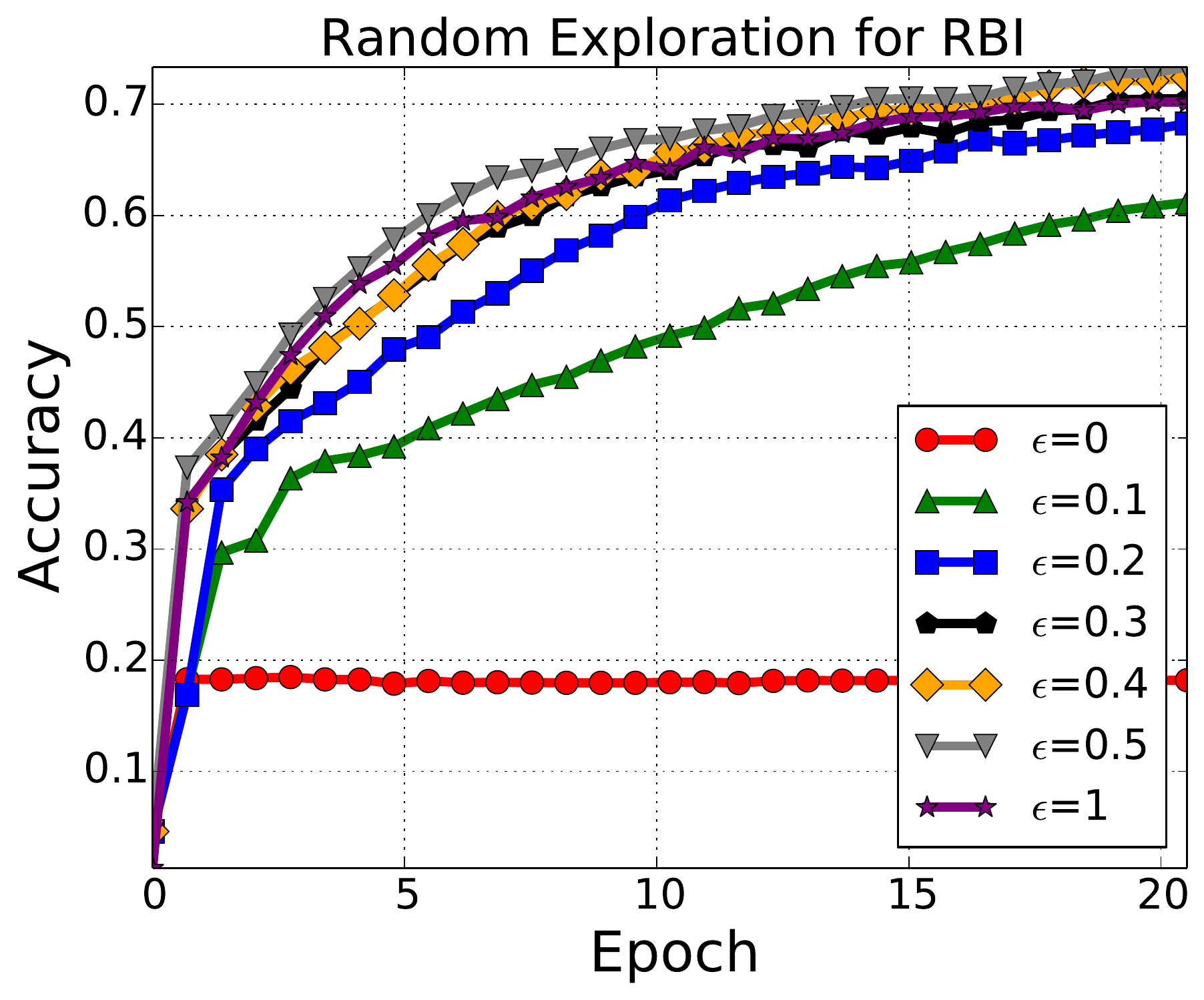}
\includegraphics[width=2.35in]{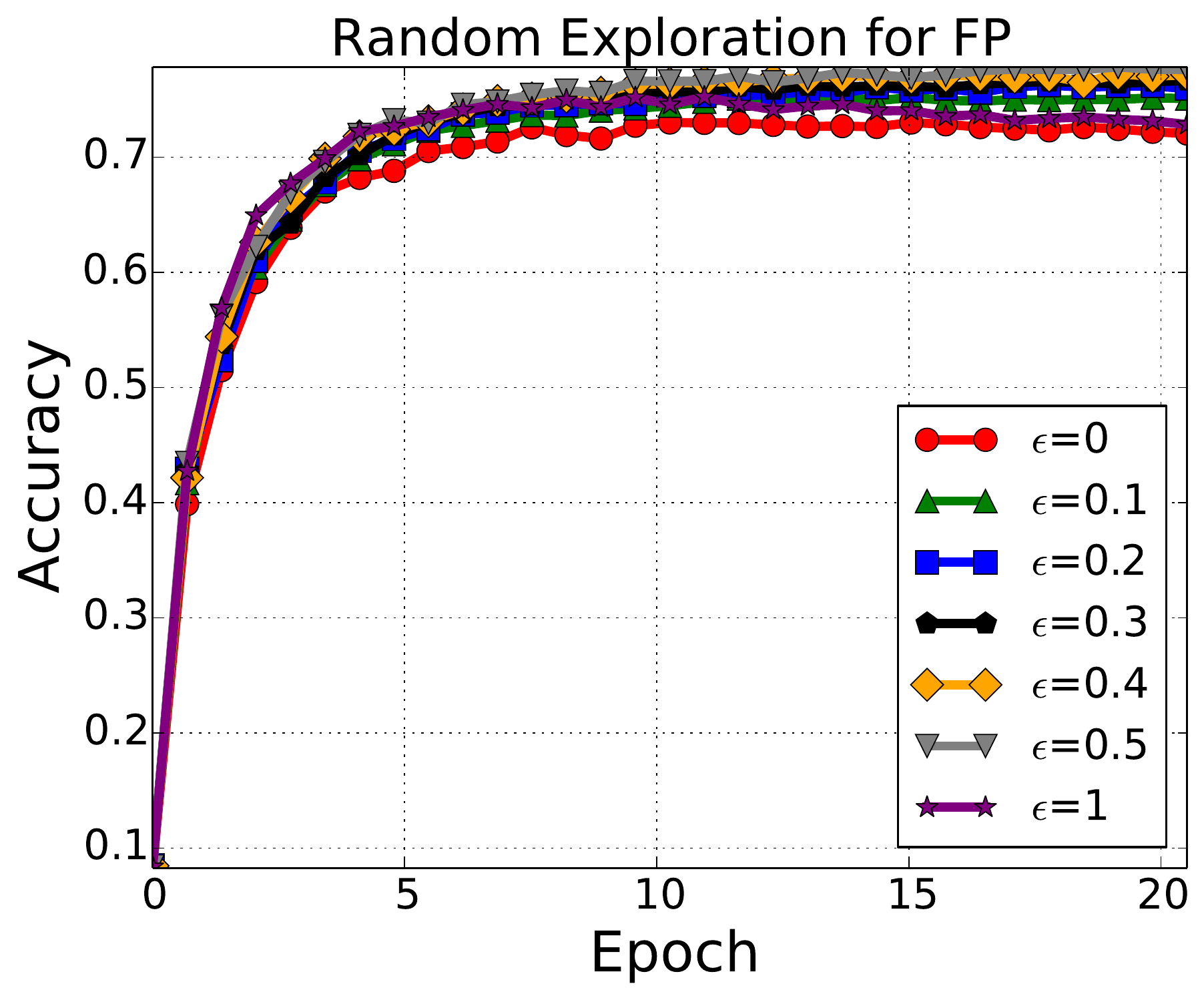}\\
\includegraphics[width=2.35in]{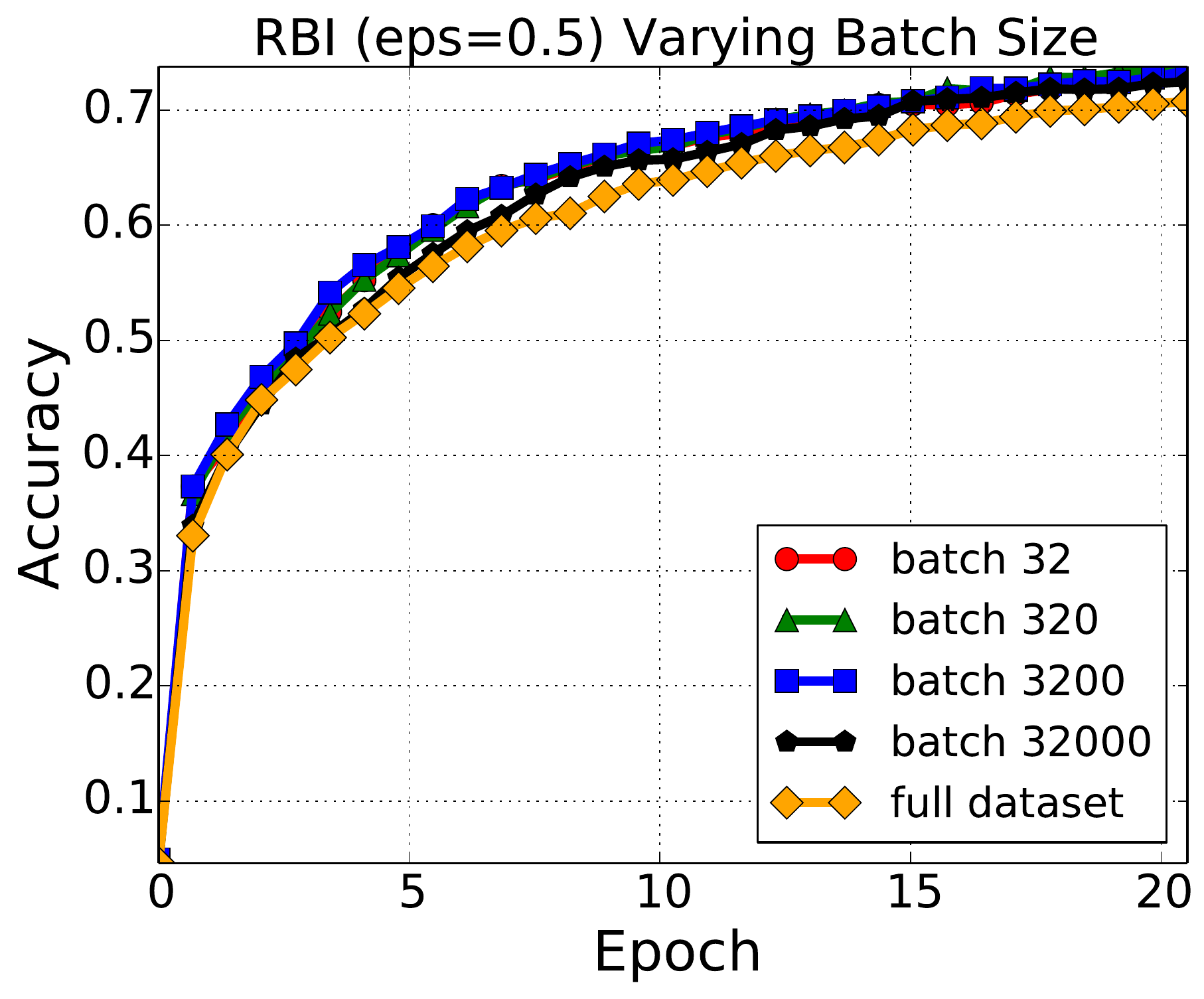}
\includegraphics[width=2.35in]{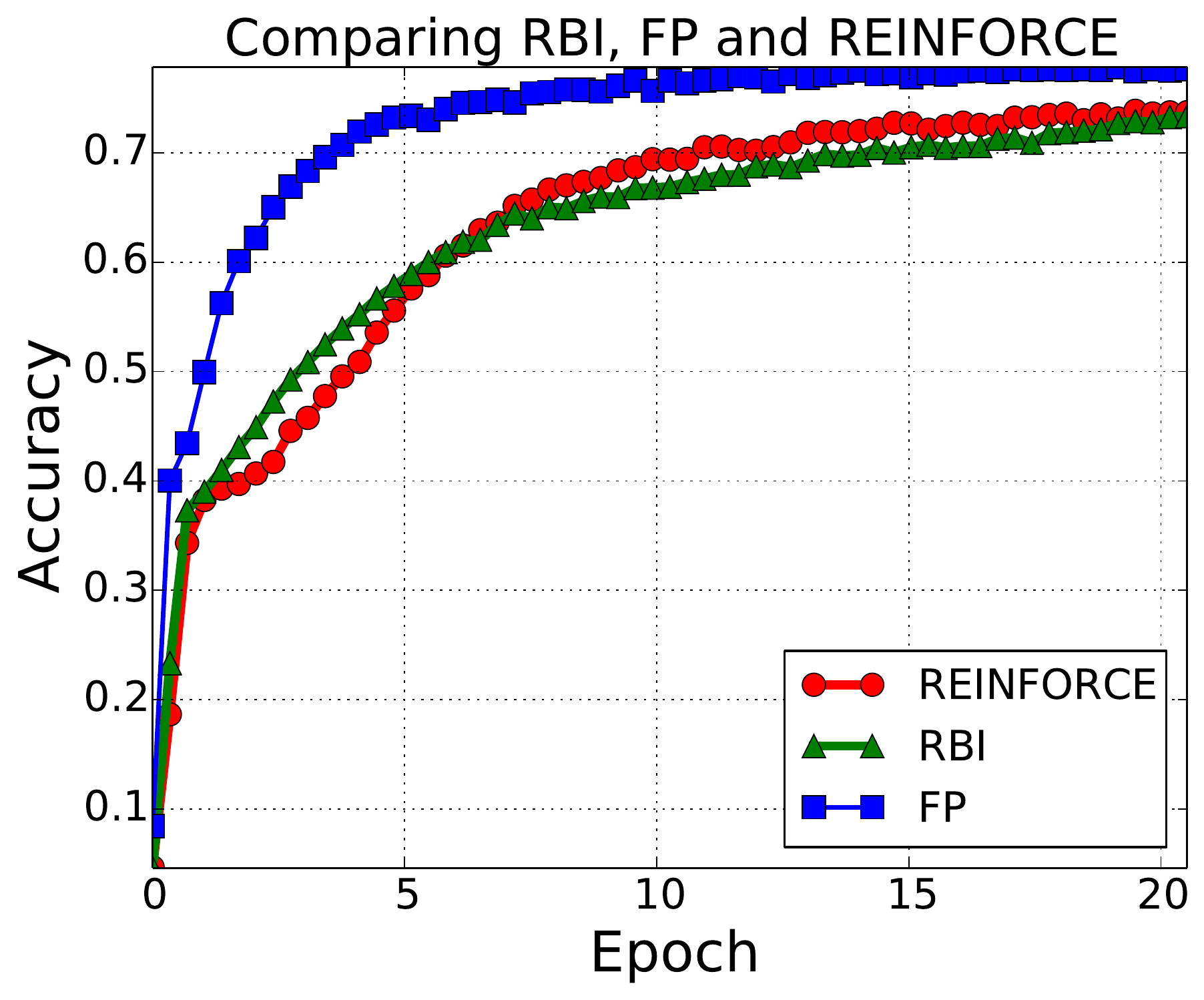}\\
\caption{{\bf WikiMovies}: Training epoch vs. test accuracy on Task $6$ varying (top left panel) exploration rate $\epsilon$ while setting batch size to $32$ for RBI,
(top right panel) for FP,
(bottom left) batch size for RBI, and (bottom right) comparing RBI, REINFORCE and FP
  with $\epsilon=0.5$. 
The model is robust to the choice of batch size. RBI and REINFORCE perform comparably.
Note that supervised, rather than reinforcement learning, with gold standard labels
achieves 80\% accuracy on this task \citep{weston2016dialog}.
\label{fig:online-movieqa-task6}
}
\end{figure*}

In our first experiments, we considered both the bAbI and WikiMovies tasks
and varied batch size, random exploration rate $\epsilon$, and type of model.
Figure~\ref{fig:online-babi-task6} and Figure~\ref{fig:online-movieqa-task6}
shows (Task 6) results on bAbI and WikiMovies.
 Other tasks yield similar conclusions and are reported in the appendix.

Overall, we obtain the following conclusions:
\begin{itemize}
\item In general RBI and FP do work in a reinforcement learning setting, but can perform better with random exploration.
\item In particular RBI can fail without exploration. RBI needs random noise for exploring labels otherwise it can get stuck predicting a subset of labels and fail.
\item REINFORCE obtains similar performance to RBI with optimal $\epsilon$. 
\item FP with balancing or with exploration via $\epsilon$ both outperform FP alone.
\item For both RBI and FP, performance is largely independent of online batch size.
\end{itemize}

\paragraph{Dataset Batch Size Experiments}
Given that larger online batch sizes appear to work well, and that this could be important in
a real-world data collection setup where the same model is deployed to gather a large amount
 of feedback from humans,
we conducted further experiments where the batch size is exactly equal to the dataset size
 and for each batch training is completed to convergence.
After the model has been trained on the dataset,
it is deployed to collect a new dataset of questions and answers, and the process is repeated.
Table~\ref{table:dataset-batch-babi} reports test error at each iteration of training,
using the bAbI Task $6$ as the case study (see the appendix for results on other tasks).
The following conclusions can be made for this setting:
\begin{itemize}
\item RBI improves in performance as we iterate. Unlike in the online case,
RBI does not need random exploration.
We believe this is because the first batch, which is collected with a randomly initialized model,
contains enough variety of examples with positive rewards
 that the model does not get stuck predicting a subset of labels.
\item FP is not stable in this setting. 
This is because once the model gets very good
at making predictions (at the third iteration), it is not exposed to a sufficient number of
negative responses anymore.
From that point on, learning degenerates and performance drops as the model always predicts the same responses.
At the next iteration, it will recover again since it has a more balanced training set,
but then it will collapse again in an oscillating behavior.
\item FP does work if extended with balancing or random exploration with sufficiently large $\epsilon$.
\item RBI+FP also works well and helps with the instability of FP, alleviating the need for random exploration and data balancing.
\end{itemize}

Overall, our simulation
 results indicate that while a bot can be effectively trained fully online from bot-teacher interactions,
collecting real dialogue data in batches (which is easier to collect and iterate experiments over) is also a viable approach. We hence pursue the latter approach in our next set of experiments.


\begin{table*}[!tbh]
\center
\begin{tabular}{l|c|c|c|c|c|c|c}
Iteration  &     1     & 2     & 3     & 4     & 5     & 6 \\
\hline
Imitation Learning               & 0.24  & 0.23  & 0.23 & 0.22 & 0.23 & 0.23 \\
Reward Based Imitation (RBI)     & 0.74  & 0.87  & 0.90 & {\bf 0.96} & {\bf 0.96} & {\bf 0.98}  \\
Forward Pred. (FP)               & {\bf 0.99}  & {\bf 0.96}  & {\bf 1.00} & 0.30 & {\bf 1.00} & 0.29  \\
RBI+FP                           & {\bf 0.99}  & {\bf 0.96}  &  {\bf 0.97} & {\bf 0.95} & 0.94 & {\bf 0.97} \\
\hline
FP (balanced)                    & {\bf 0.99} & {\bf 0.97} & {\bf 0.97} & {\bf 0.97} & {\bf 0.97} & {\bf 0.97} \\
FP (rand. exploration $\epsilon=0.25$)                & {\bf {\bf 0.96}} & 0.88 & 0.94 & 0.26 & 0.64 & {\bf 0.99} \\
FP (rand. exploration $\epsilon=0.5$)                 & {\bf 0.98} & {\bf 0.98} & {\bf 0.99} & {\bf 0.98} & {\bf 0.95} & {\bf 0.99} \\
\end{tabular}
\caption{Test accuracy of various models per iteration in the dataset batch size case
(using batch size equal to the size of the full training set) for bAbI, Task $6$.
 Results $>0.95$ are in bold.
\label{table:dataset-batch-babi}
}
\end{table*}

\paragraph{Relation to experiments in \cite{weston2016dialog}}
As described in detail in Section \ref{sec:related} the datasets we use in our experiments were
introduced in \citep{weston2015towards}. However, that work involved
 constructing pre-built fixed policies (and hence, datasets),
rather than training the learner in a reinforcement/interactive learning using
 a simulator, as in our work. 
They achieved this by choosing an omniscient (but deliberately imperfect) labeler that gets $\pi_{acc}$
examples always correct (the paper looked at values 1\%, 10\% and 50\%).
 In a realistic setting one does not have access to an omniscient labeler, one has to
 learn a policy completely from scratch, online, starting with a random policy, as we do here.
Nevertheless, it is possible to compare our {\em learnt} policies to those
results because we use the same train/valid/test splits.

The clearest comparison comparison is via Table \ref{table:dataset-batch-babi},
where the policy is learnt using batch iterations of the dataset, updating the policy on each
iteration. \cite{weston2015towards}  can be viewed as training only one iteration, with a pre-built policy, 
as explained above, where 59\%, 81\% and 99\% accuracy was obtained for RBI for $\pi_{acc}$ with
1\%, 10\% and 50\% respectively\footnote{Note, this is not the same as a randomly initialized neural network 
policy, because due to the synthetic construction with an omniscient labeler the labels will be balanced. In our work, we learn the policy from randomly initialized weights which are updated as we learn the policy.}. While $\pi_{acc}$ of 50\% is good enough to solve the task, lower values
are not. In this work a random policy begins with 74\% accuracy on the first iteration, but importantly
on each iteration the policy is updated and improves, with values of 87\%, 90\% on iterations 2 and 3 respectively, and 98\% on iteration 6.
This is a key differentiator to the work of  \citep{weston2015towards} where such improvement was not shown.
We show that such online learning works for both reward-based numerical feedback and for forward prediction methods using textual feedback (as long as balancing or random exploration is performed sufficiently). The final performance outperforms most values of  $\pi_{acc}$ from \cite{weston2015towards} unless $\pi$ is so large that the task is already solved. This is a key contribution of our work.

Similar conclusions can be made for Figures \ref{fig:online-babi-task6} and \ref{fig:online-movieqa-task6}.
Despite our initial random policy starting at close
to 0\% accuracy, if random exploration $\epsilon \ge 0.2$ is employed then after a number of  epochs the performance is better than most values of $\pi_{acc}$ from \cite{weston2015towards}, e.g. compare the accuracies given in the previous paragraph (59\%, 81\% and 99\%) to Figure \ref{fig:online-babi-task6}, top left.

\subsection{Human Feedback} \label{sec:mturkexp}
We employed Turkers to both ask questions and then give textual feedback
on the bot's answers, as described in Section \ref{sec:data-mturk}.
Our experimental protocol was as follows.
We first trained a MemN2N using supervised (i.e., imitation) learning on a training 
set of 1000 questions produced by Turkers and using the known correct answers 
provided by the original dataset (and no textual feedback).
Next, using the trained policy, we collected textual feedback for the responses
of the bot for an additional 10,000 questions.
Examples from the collected dataset are given in Figure \ref{fig:mturk_data}.
Given this dataset, we compare various models: RBI, FP and FP+RBI.
As we know the correct answers to the additional questions, we can  assign a positive reward
to questions the bot got correct. We hence measure the impact of the sparseness of this
reward signal, where a fraction $r$ of additional examples have rewards.
The models are tested on a test set of $\sim$8,000 questions (produced by Turkers),
and hyperparameters are tuned on a similarly sized validation set.
Note this is a harder task than the WikiMovies task in the simulator due to 
the use natural language from Turkers, hence lower test performance is expected.

Results are given in Table \ref{table:mturk-res}.
They indicate that both RBI and FP are useful.
When rewards are sparse, FP still works via the textual feedback 
while RBI can only use the initial 1000 examples when $r=0$.
As FP does not use numericalrewards at all, it is invariant to the parameter $r$.
The combination of FP and RBI outperforms either alone.

\begin{table*}[!tbh]
\begin{center}
\begin{tabular}{l|c|c|c|c|}
Model                         & $r=0$   &  $r=0.1$  &  $r=0.5$  & $r=1$ \\
\hline
Reward Based Imitation (RBI)  & 0.333    &  0.340   &  0.365   & 0.375 \\
Forward Prediction (FP)       & 0.358    &  0.358   &  0.358   & 0.358 \\
RBI+FP                        & 0.431    &  0.438   &  0.443   & 0.441 \\
\end{tabular}
\end{center}
\caption{{\bf Incorporating Feedback From Humans via Mechanical Turk.}
Textual feedback is provided for 10,000 model predictions (from a model trained with 1k labeled
 training examples), and additional sparse
binary rewards (fraction $r$ of examples have rewards).
Forward Prediction
and Reward-based Imitation are both useful, with their combination performing best.
\label{table:mturk-res}
}
\end{table*}

We also conducted
additional experiments comparing with (i) synthetic feedback and (ii) the fully supervised
case which are given in Appendix \ref{sec:appendix-mturk}.
They show that the results with human feedback are competitive with these approaches.

%% file: conc.tex
We studied dialogue learning of end-to-end models using textual feedback and numerical rewards.
Both fully online and iterative batch settings are viable approaches to policy learning, as long as
possible instabilities in the learning algorithms are taken into account. Secondly, we showed 
for the first time that the recently introduced FP method can work
in both an online setting and on real human feedback.
Overall, our results indicate that it is feasible to build a practical
 pipeline that starts with a model trained on an initial fixed dataset,
which then learns from interactions with humans  in a (semi-)online fashion to improve itself.
Future research should work towards doing this in a never-ending learning setup.



%% file: extra_data.tex
The tasks in  \cite{weston2016dialog} were specifically:\\
- {\bf   Task 1}: The teacher tells the student exactly what they should have said (supervised baseline).\\
- {\bf  Task 2}:
The teacher replies with positive textual feedback and reward, or negative textual feedback. \\
- {\bf Task 3}: The teacher gives textual feedback containing the answer when the bot is wrong.\\
- {\bf Task 4}: The teacher provides a hint by providing the class of the correct answer, e.g., ``No it's a movie" for the question ``which movie did Forest Gump star in?".\\
- {\bf Task 5}: The teacher provides
a reason why the student's answer is wrong by pointing out the relevant supporting fact from the knowledge base.\\
- {\bf Task 6}: The teacher gives positive reward only 50\% of the time. \\
- {\bf Task 7}: Rewards are missing and the teacher only gives natural language feedback.\\
- {\bf Task 8}: Combines Tasks 1 and 2 to see whether a learner can learn successfully from both forms of supervision at once.\\
- {\bf Task 9}:  The bot asks questions of the teacher about what it has done wrong.\\
- {\bf Task 10}:  The bot will receive a hint rather than the correct answer after asking for help.

We refer the readers to \citep{weston2016dialog} for more detailed descriptions and
the motivation behind these tasks.
The difference in our system is that the model can be trained on-the-fly via the simulator:
after receiving feedback and/or rewards, the model can update itself and apply its learning
to the next episode. We present results on Tasks 2, 3 and 4 in this appendix


\begin{figure*}[!ht]
\centering
\footnotesize
\begin{tabular}{|l|l|l|}
\cline{1-1}\cline{3-3}
 \\[-2ex]
{\color{blue}T : Which  {\color{blue} movie} did Tom Hanks {\color{blue} star} in ?} &&{\color{blue}T : Which  {\color{blue}movie} did Tom Hanks  {\color{blue}star} in ?}  \\
{\color{red}S : Forrest Gump }  &&  {\color{red}S :  Brad Pitt.}\\
\cline{1-1}\cline{3-3}
 \multicolumn{1}{c}{}
\vspace{-2mm} \\
\cline{1-1}\cline{3-3}
Task 1: Imitating an Expert Student &&Task 1: Imitating an Expert Student  \\
{\color{blue}S: Forrest Gump} && {\color{blue}S: Forrest Gump}\\
{\color{blue}T: (no response)  } &&{\color{blue}T: (no response)  }  \\
\cline{1-1}\cline{3-3}
 \multicolumn{1}{c}{}
\vspace{-2mm} \\
\cline{1-1}\cline{3-3}
Task 2: Positive and Negative Feedback &&Task 2: Positive and Negative Feedback  \\
{\color{blue}T: Yes, that's right! (+)} && {\color{blue}T: No, that's incorrect! \MINUS} \\
\cline{1-1}\cline{3-3}
 \multicolumn{1}{c}{}
\vspace{-2mm} \\
\cline{1-1}\cline{3-3}
Task 3: Answers Supplied by Teacher &&Task 3: Answers Supplied by Teacher  \\
{\color{blue}T: Yes, that is correct. (+)} && {\color{blue}T: No, the answer is Forrest Gump ! \MINUS} \\
\cline{1-1}\cline{3-3}
 \multicolumn{1}{c}{}
\vspace{-2mm} \\
\cline{1-1}\cline{3-3}
Task 4: Hints Supplied by Teacher &&Task 4: Hints Supplied by Teacher  \\
{\color{blue}T: Correct! (+)} && {\color{blue}T: No, it's a movie ! \MINUS} \\
\cline{1-1}\cline{3-3}
 \multicolumn{1}{c}{}
\vspace{-2mm} \\
\cline{1-1}\cline{3-3}
Task 5: Supporting Facts Supplied by Teacher &&Task 5: Supporting Facts Supplied by Teacher  \\
{\color{blue}T: That's right. (+)} && {\color{blue}T: No, because Forrest Gump starred actors } \\
&&{\color{blue} Tom Hanks, Robin Wright, Gary Sinise ! \MINUS} \\
\cline{1-1}\cline{3-3}
 \multicolumn{1}{c}{}
\vspace{-2mm} \\
\cline{1-1}\cline{3-3}
Task 6: Partial Feedback && Task 6: Partial Feedback \\
if random(0,1)$<$0.5 then  && {\color{blue}T: Sorry, wrong. \MINUS} \\
~~~~{\color{blue}T: That's correct. (+)} && \\
else ~{\color{blue}T: That's correct.  } && \\
\cline{1-1}\cline{3-3}
 \multicolumn{1}{c}{}
\vspace{-2mm} \\
\cline{1-1}\cline{3-3}
Task 7: No Feedback  && Task 7: No Feedback \\
{\color{blue}T: Yes.} && {\color{blue}T: No.} \\
\cline{1-1}\cline{3-3}
 \multicolumn{1}{c}{}
\vspace{-2mm} \\
\cline{1-1}\cline{3-3}
Task 8: Imitation and Feedback Mixture &&Task 8: Imitation and Feedback Mixture \\
if random(0,1)$<$0.5 then  && if random(0,1)$<$0.5 then  \\
~~~~{\color{blue}T: Yes, that's right! (+)} &&~~~~ {\color{blue}T: Wrong. \MINUS} \\
else ~{\color{blue}T: (no response)  } && else ~{\color{blue} S: Forrest Gump  } \\
\cline{1-1}\cline{3-3}
 \multicolumn{1}{c}{}
\vspace{-2mm} \\
\cline{1-1}\cline{3-3}
Task 9: Asking For Corrections &&Task 9: Asking For Corrections  \\
{\color{blue}T: Correct! (+)}  &&  {\color{blue}T: No, that's wrong. } \\
&& {\color{red}S: Can you help me? } \\
&& {\color{blue}T: Forrest Gump !} \\
\cline{1-1}\cline{3-3}
 \multicolumn{1}{c}{}
\vspace{-2mm} \\
\cline{1-1}\cline{3-3}
Task 10: Asking For Supporting Facts && Task 10: Asking For Supporting Facts\\
{\color{blue}T: Yes, that's right! (+)}  &&  {\color{blue}T: Sorry, that's not it.} \\
&& {\color{red}S: Can you help me? } \\
&& {\color{blue}T: A relevant fact is that
Forrest Gump starred actors } \\
&&{\color{blue} Tom Hanks, Robin Wright, Gary Sinise ! \MINUS} \\\hline

\end{tabular}
\caption{The ten tasks our simulator implements, which evaluate different
forms of teacher response and binary feedback.
In each case the same example from WikiMovies is given for simplicity,
where the student answered correctly for all tasks (left) or incorrectly (right).
 {\color{red}Red} text denotes responses by the bot with  {\color{red}S} denoting the bot.
{\color{blue}Blue} text is spoken by the teacher with {\color{blue}T} denoting the teacher's response. For imitation learning the teacher provides the response the student should say denoted with {\color{blue}S} in Tasks 1 and 8.
A {(\color{blue}{+})} denotes a positive reward.
}
\label{Tasks}
\end{figure*}

%% file: mturk.tex
These are the instructions given for the textual feedback mechanical turk task
(we also constructed a separate task to collect the initial questions, not described here):\\

Title: Write brief responses to given dialogue exchanges (about 15 min)\\

Description: Write a brief response to a student's answer to a teacher's question, providing feedback to the student on their answer.

Instructions:\\

Each task consists of the following triplets:
\begin{enumerate}
\item a question by the teacher

\item the correct answer(s) to the question (separated by ``OR'')

\item a proposed answer in reply to the question from the student

\end{enumerate}

Consider the scenario where you are the teacher and have already asked the question, and received the reply from the student. Please compose a brief response giving feedback to the student about their answer. The correct answers are provided so that you know whether the student was correct or not.

For example, given 1) question: ``what is a color in the united states flag?''; 2) correct answer: ``white, blue, red''; 3) student reply: ``red'', your response could be something like ``that's right!''; for 3) reply: ``green'', you might say ``no that's not right'' or ``nope, a correct answer is actually white''.

Please vary responses and try to minimize spelling mistakes. If the same responses are copied/pasted or overused, we'll reject the HIT.

Avoid naming the student or addressing ``the class'' directly.

We will consider bonuses for higher quality responses during review.

%% file: extra_exp.tex
\begin{table*}[!h]
\center
\begin{tabular}{c|c|c|c|c|c|c}
Iteration  & 1     & 2     & 3     & 4     & 5     & 6 \\ 
\hline
Imitation Learning           & 0.24       & 0.23      & 0.23 & 0.23 & 0.25 & 0.25 \\
Reward Based Imitation (RBI) & {\bf 0.95} & {\bf 0.99} & {\bf 0.99} & {\bf 0.99} & \bf{1.00}     & {\bf 1.00 }    \\
Forward Pred. (FP)           & {\bf 1.00} & 0.19      & 0.86 & 0.30 & {\bf 99} & 0.22  \\
RBI+FP                       & {\bf 0.99} & {\bf 0.99} & {\bf 0.99} & {\bf 0.99} & {\bf 99} & {\bf 0.99} \\
\hline
FP (balanced)                          & {\bf 0.99} & {\bf 0.97} & {\bf 0.98} & {\bf 0.98} & {\bf 0.96} & {\bf 0.97} \\
FP (rand. exploration $\epsilon=0.25$)  & {\bf 0.99} & 0.91 & 0.93 & 0.88  & 0.94 & 0.94 \\ 
FP (rand. exploration $\epsilon=0.5$)   & {\bf 0.98} & 0.93  & {\bf 0.97} & {\bf 0.96} & {\bf 0.95} & {\bf 0.97} \\
\end{tabular}
\caption{Test accuracy of various models in the dataset batch size case 
(using batch size equal to the size of the full training set) for bAbI, task $3$.
 Results $>0.95$ are in bold.
\label{acc}
}
\end{table*}
\begin{figure*}[h!]
\includegraphics[width=2.5in]{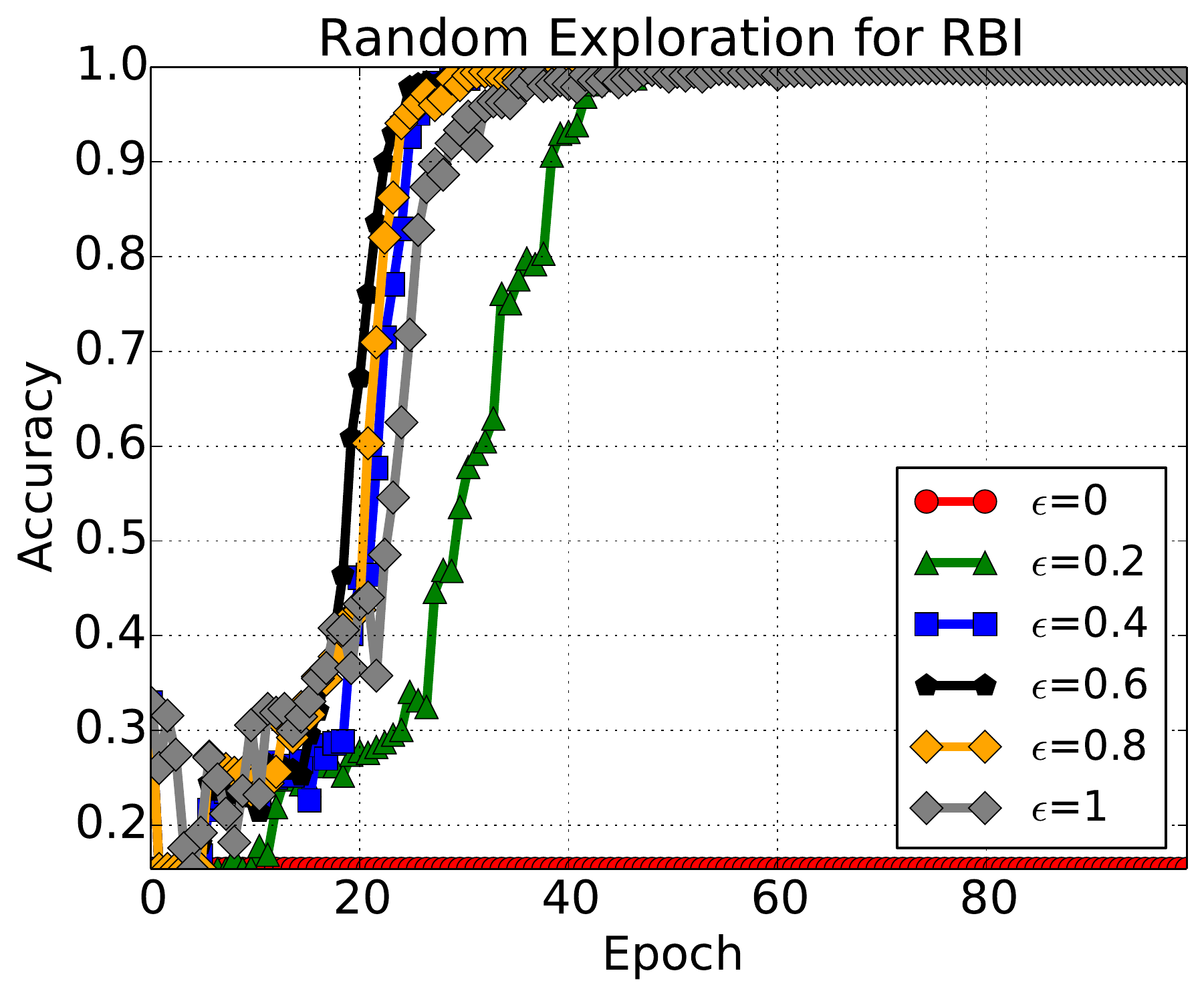}
\includegraphics[width=2.5in]{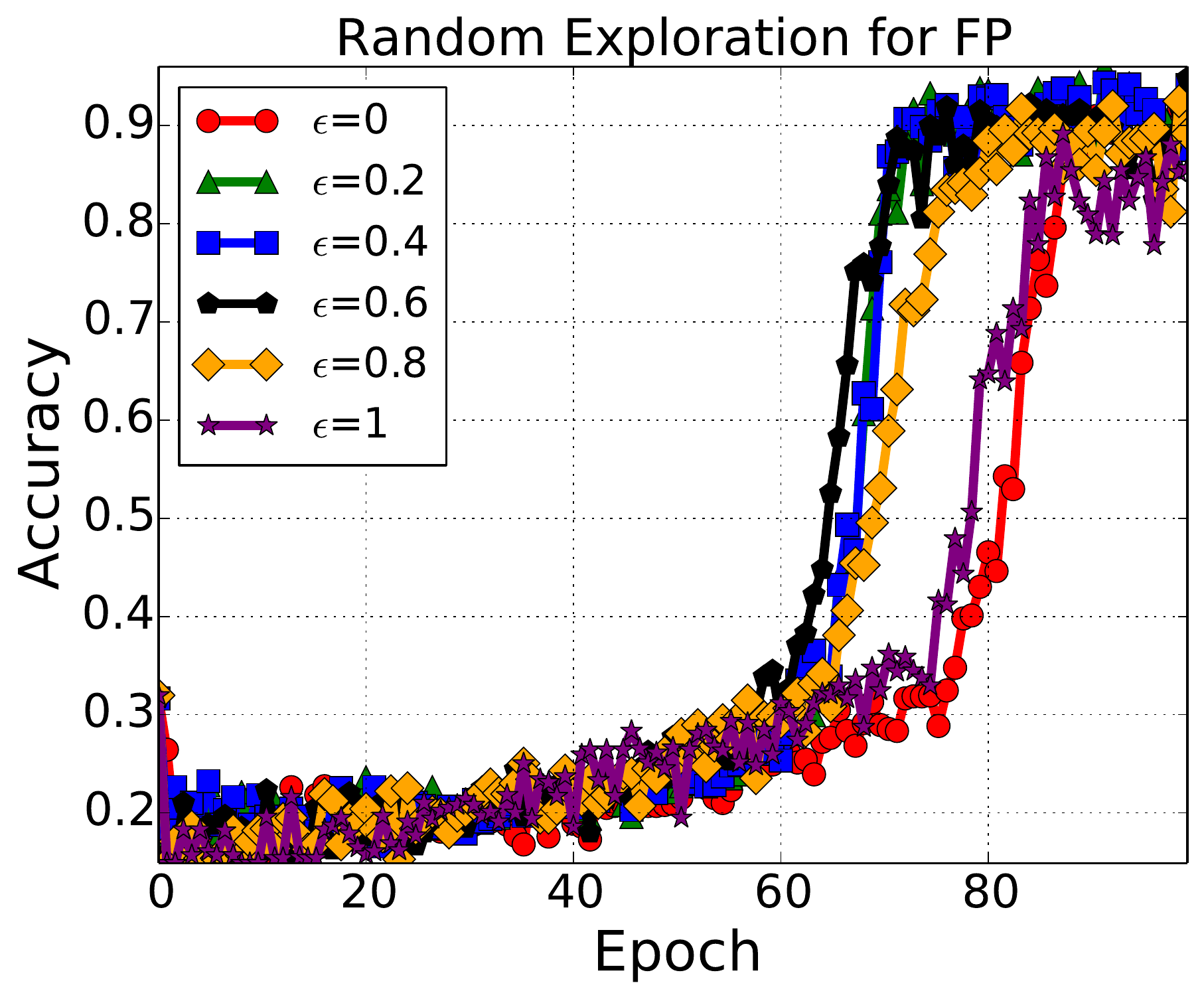}\\
\includegraphics[width=2.5in]{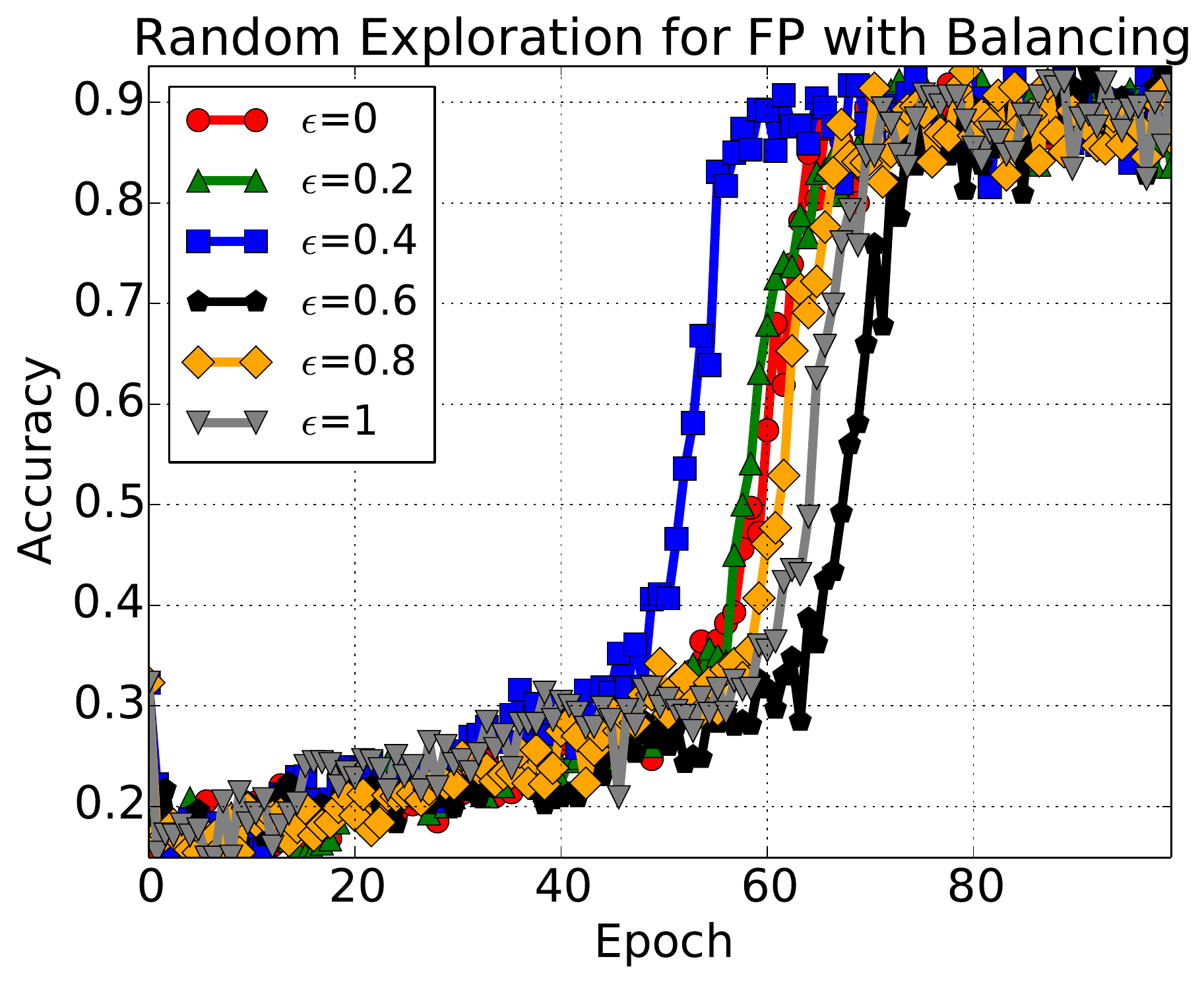}
\includegraphics[width=2.5in]{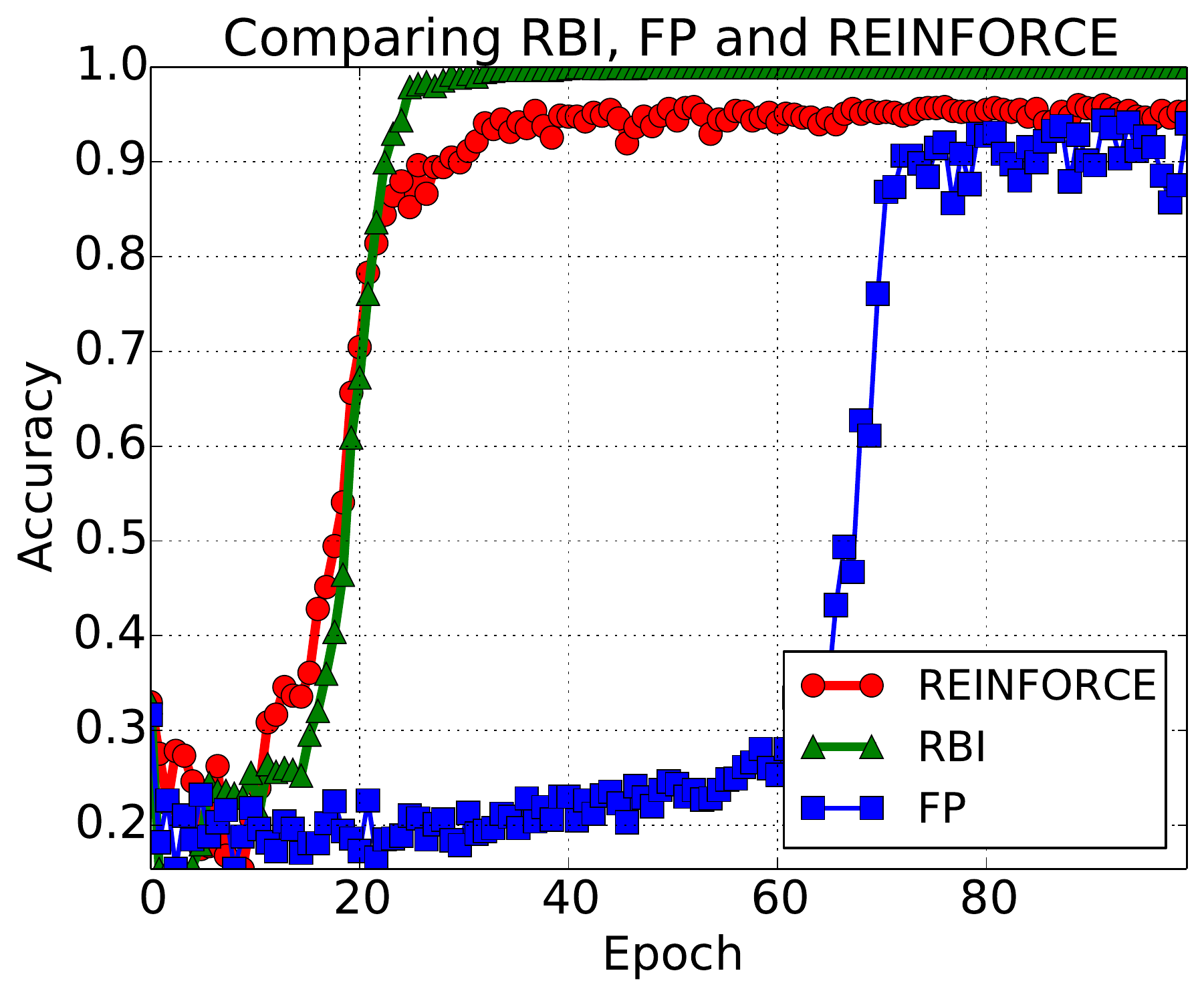}\\
\includegraphics[width=2.5in]{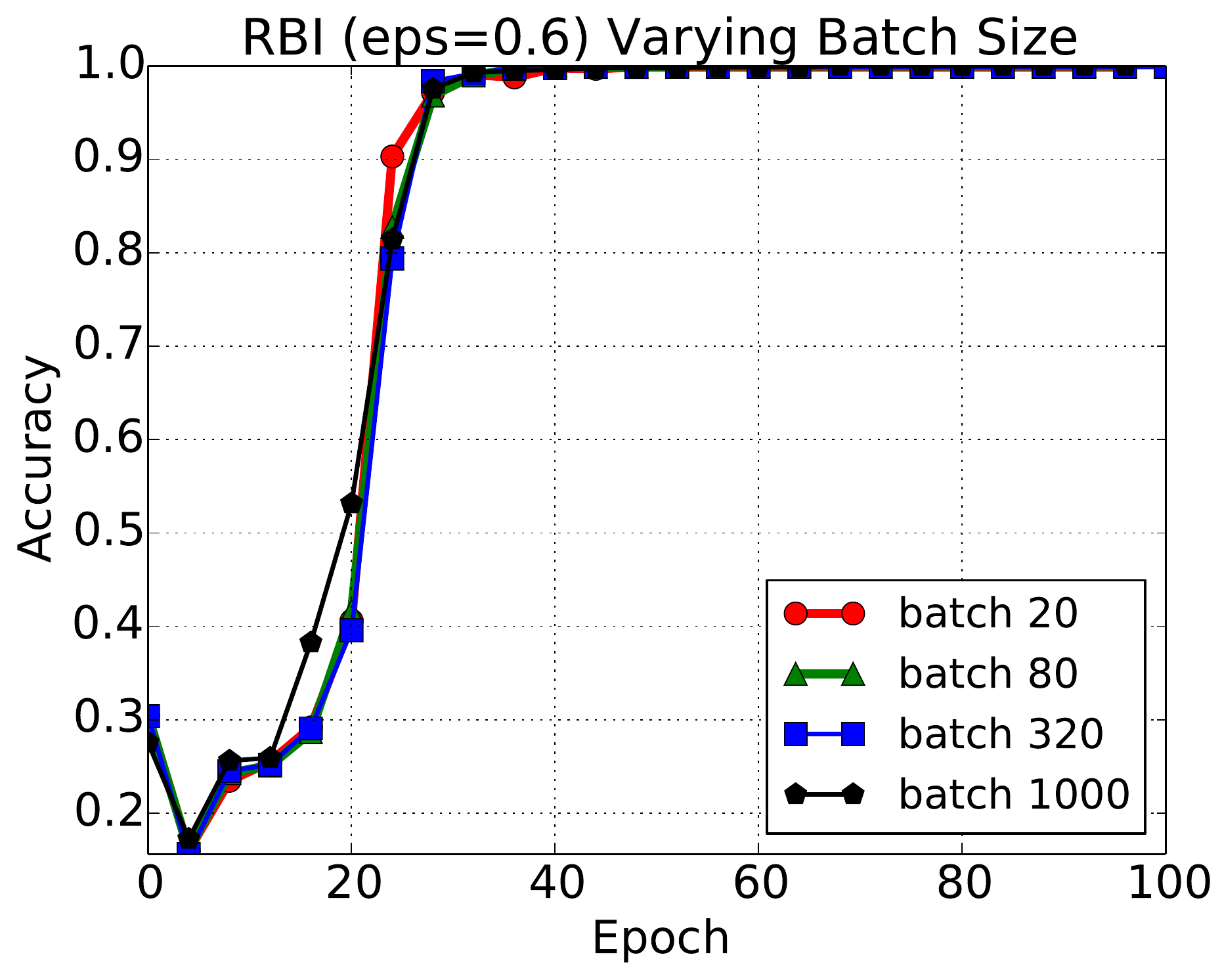}
\includegraphics[width=2.5in]{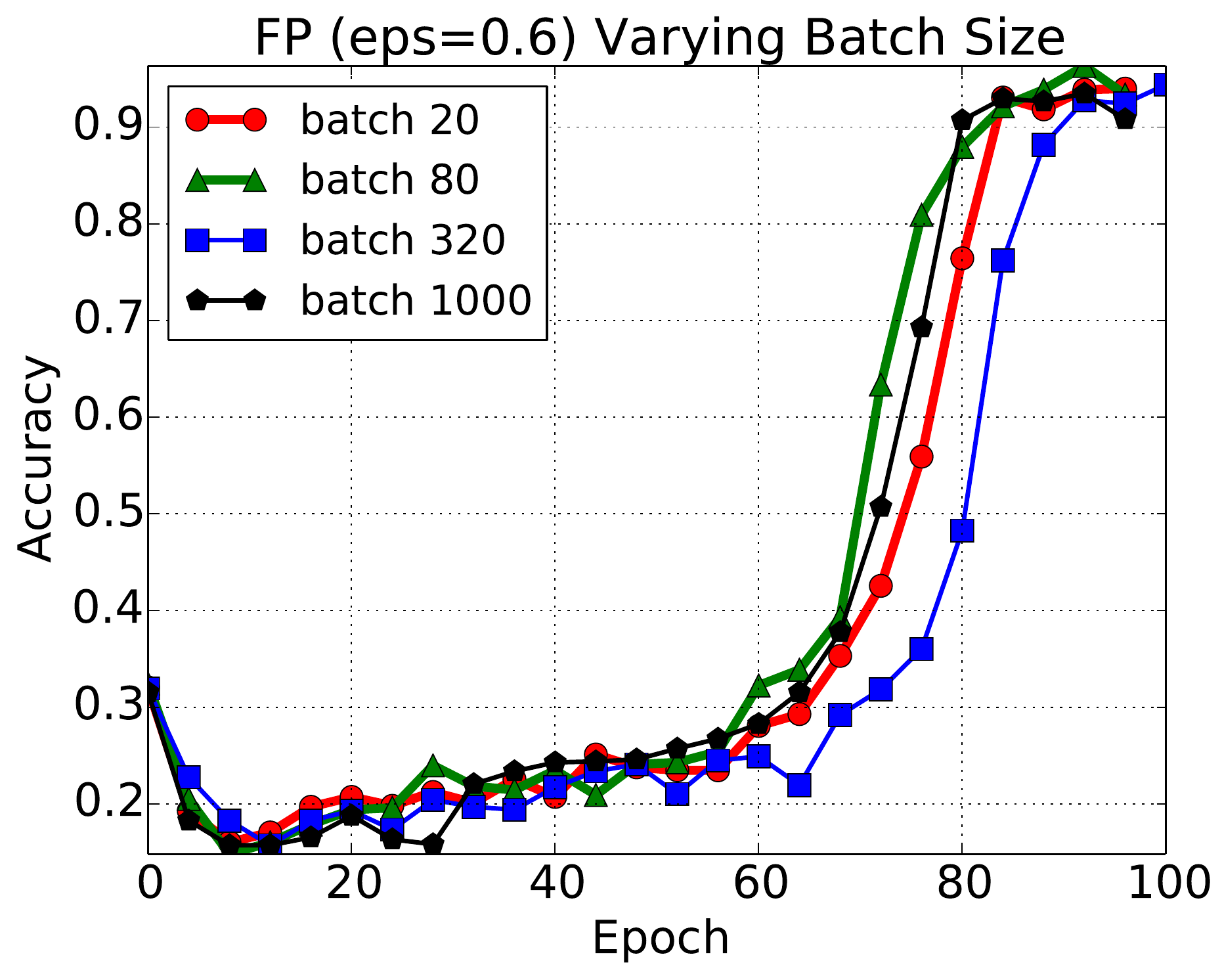}\\
\caption{{\bf  Training epoch vs. test accuracy for bAbI (Task 2) varying exploration $\epsilon$ and batch size.}
\label{fig:online-babi-task2}
}
\end{figure*}
\begin{figure*}[h!]
\includegraphics[width=2.5in]{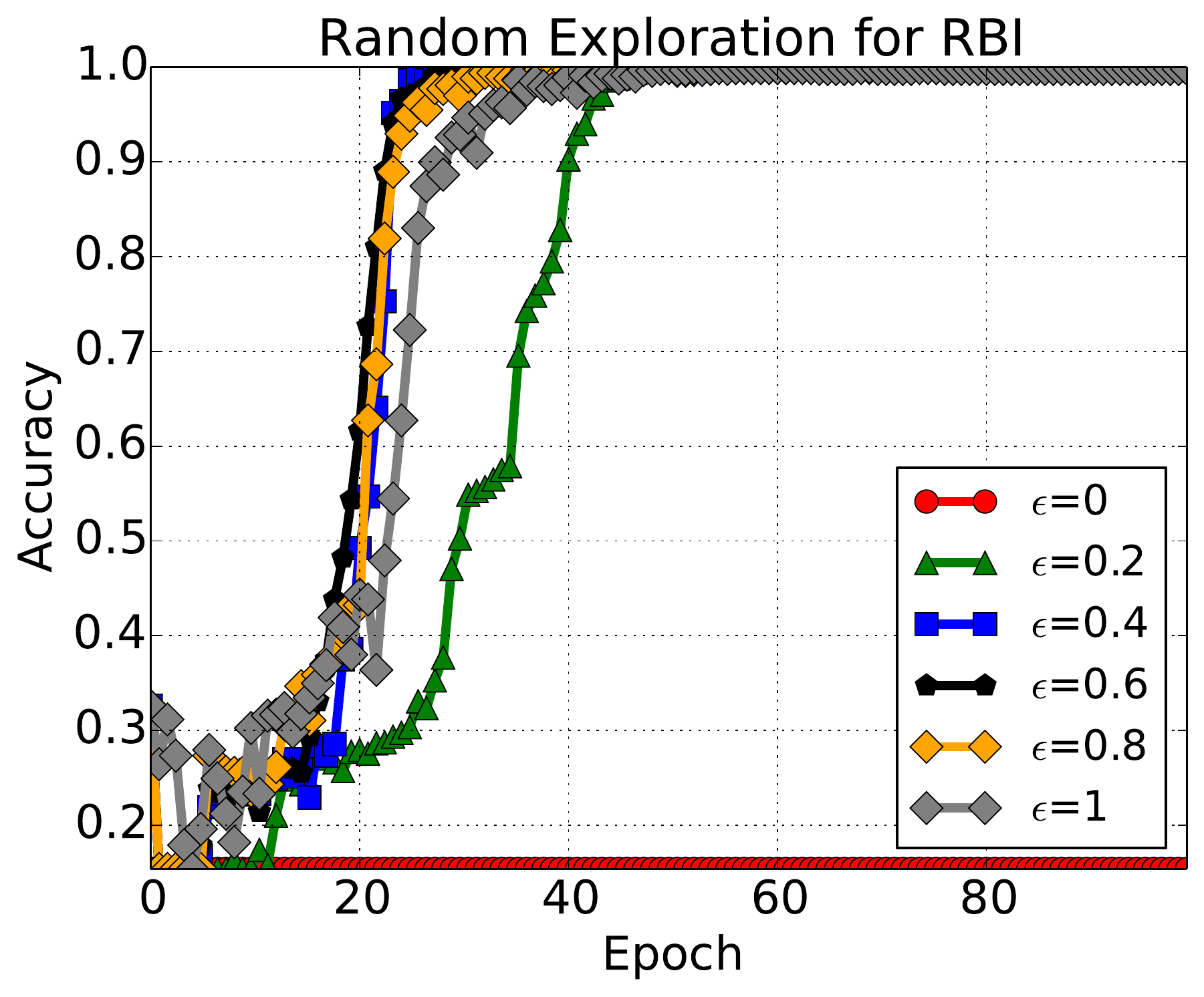}
\includegraphics[width=2.5in]{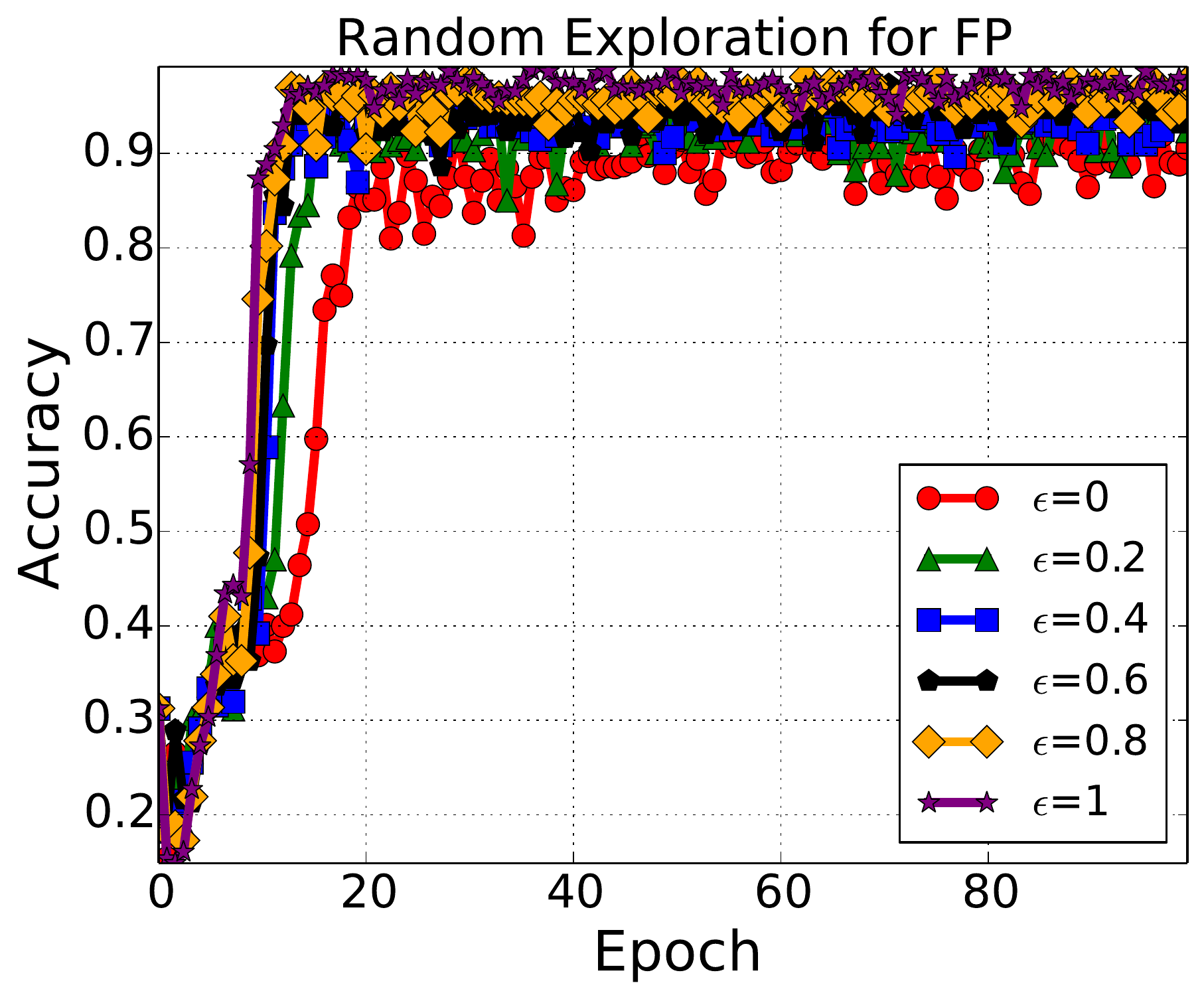}\\
\includegraphics[width=2.5in]{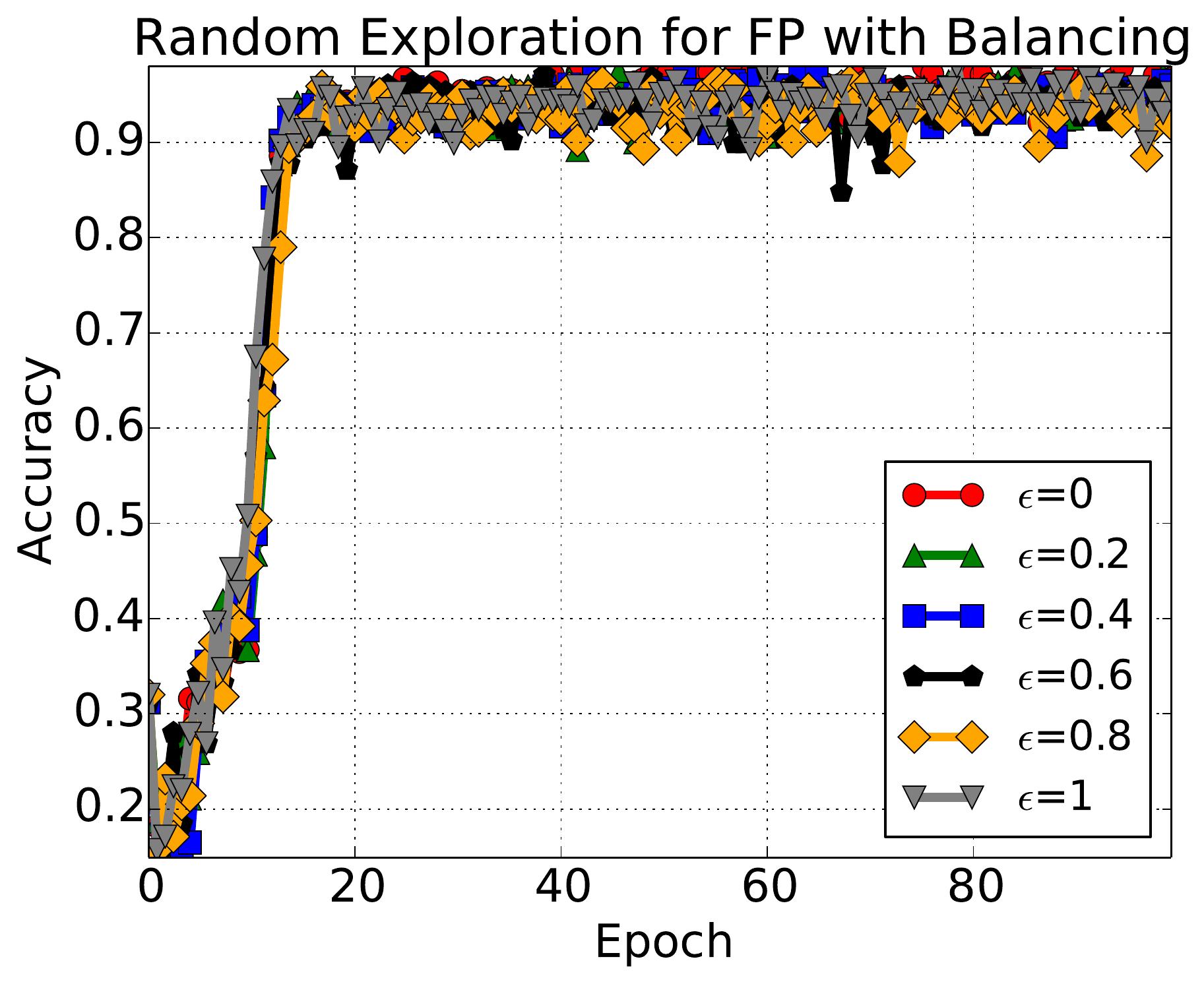}
\includegraphics[width=2.5in]{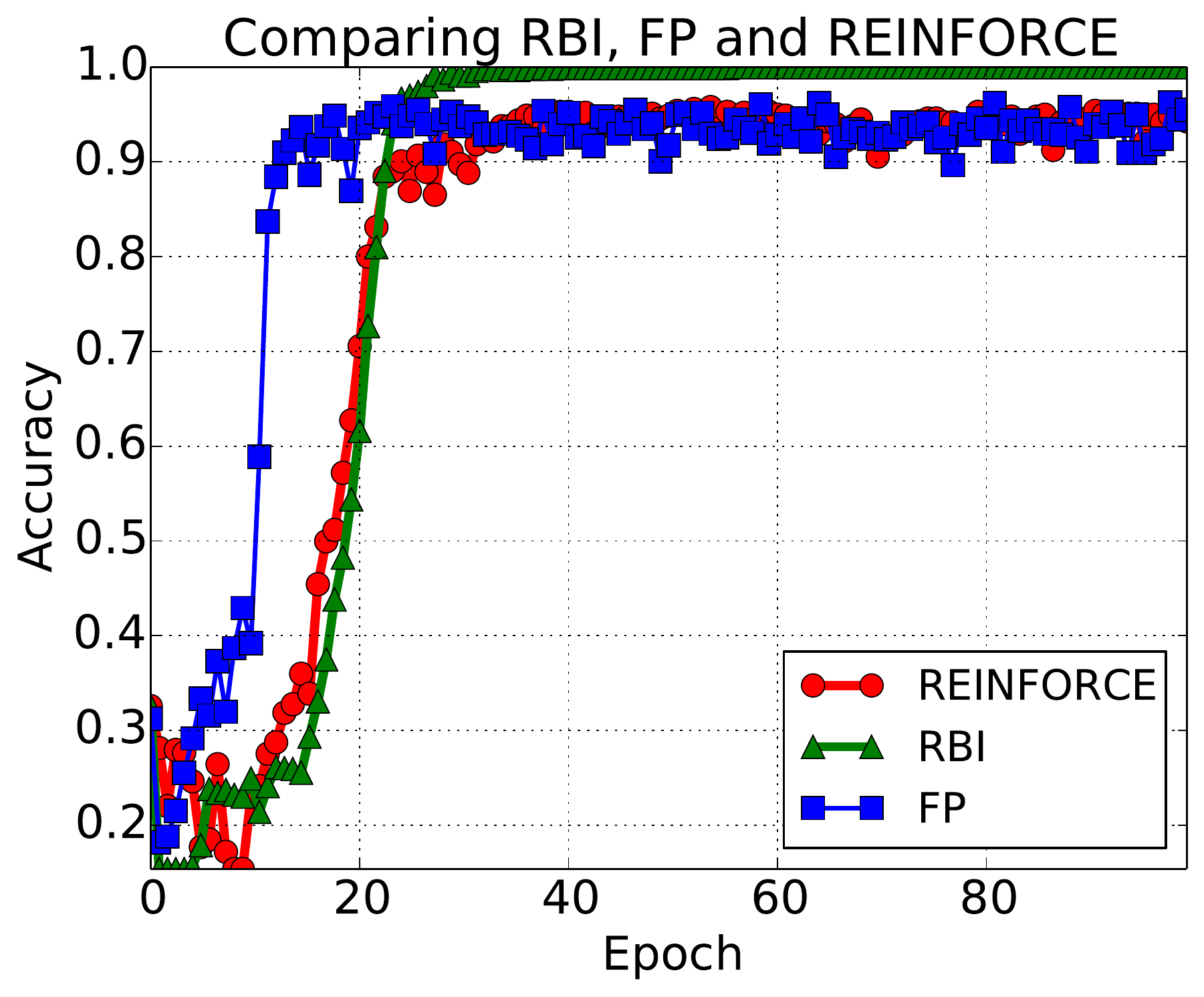}\\
\includegraphics[width=2.5in]{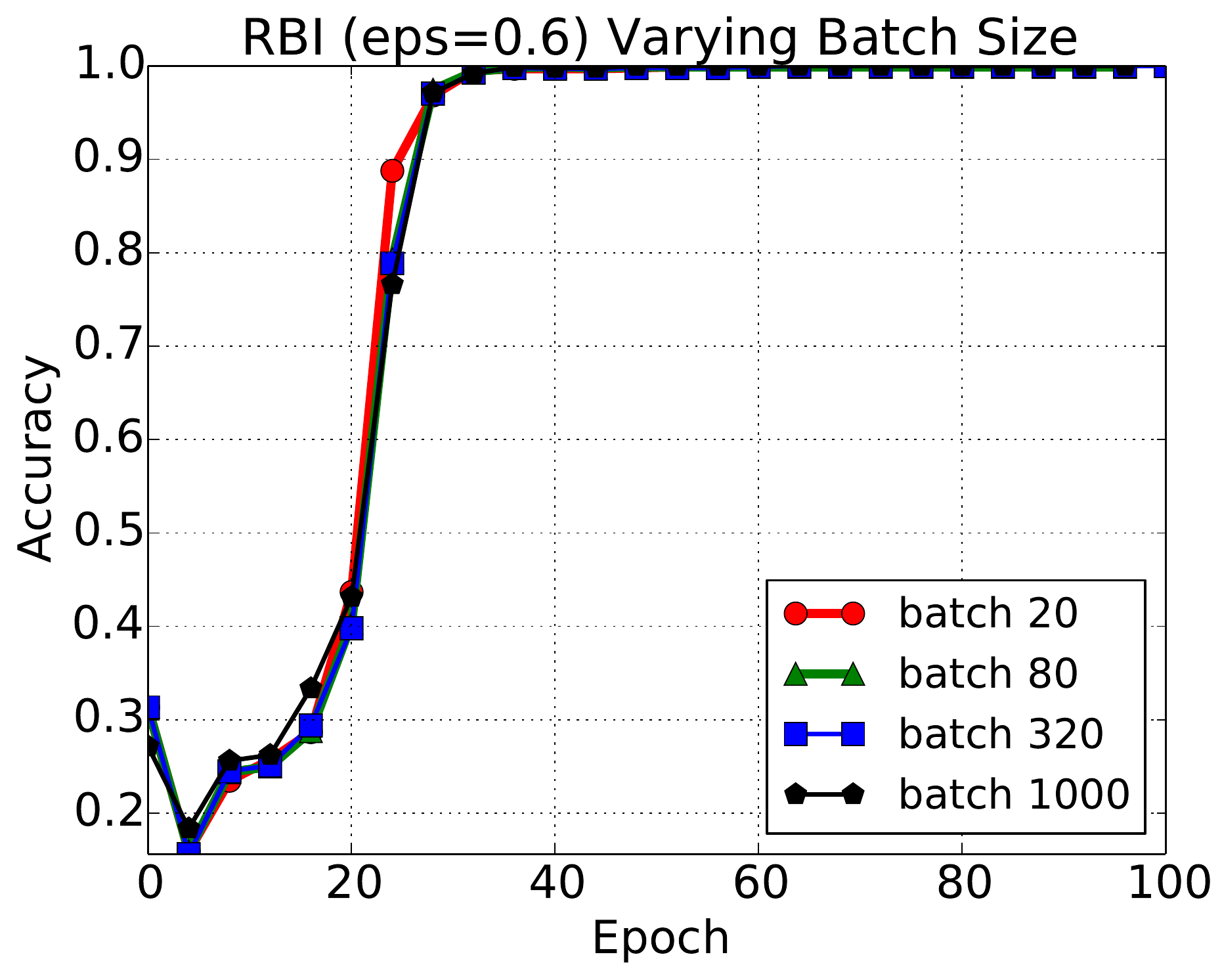}
\includegraphics[width=2.5in]{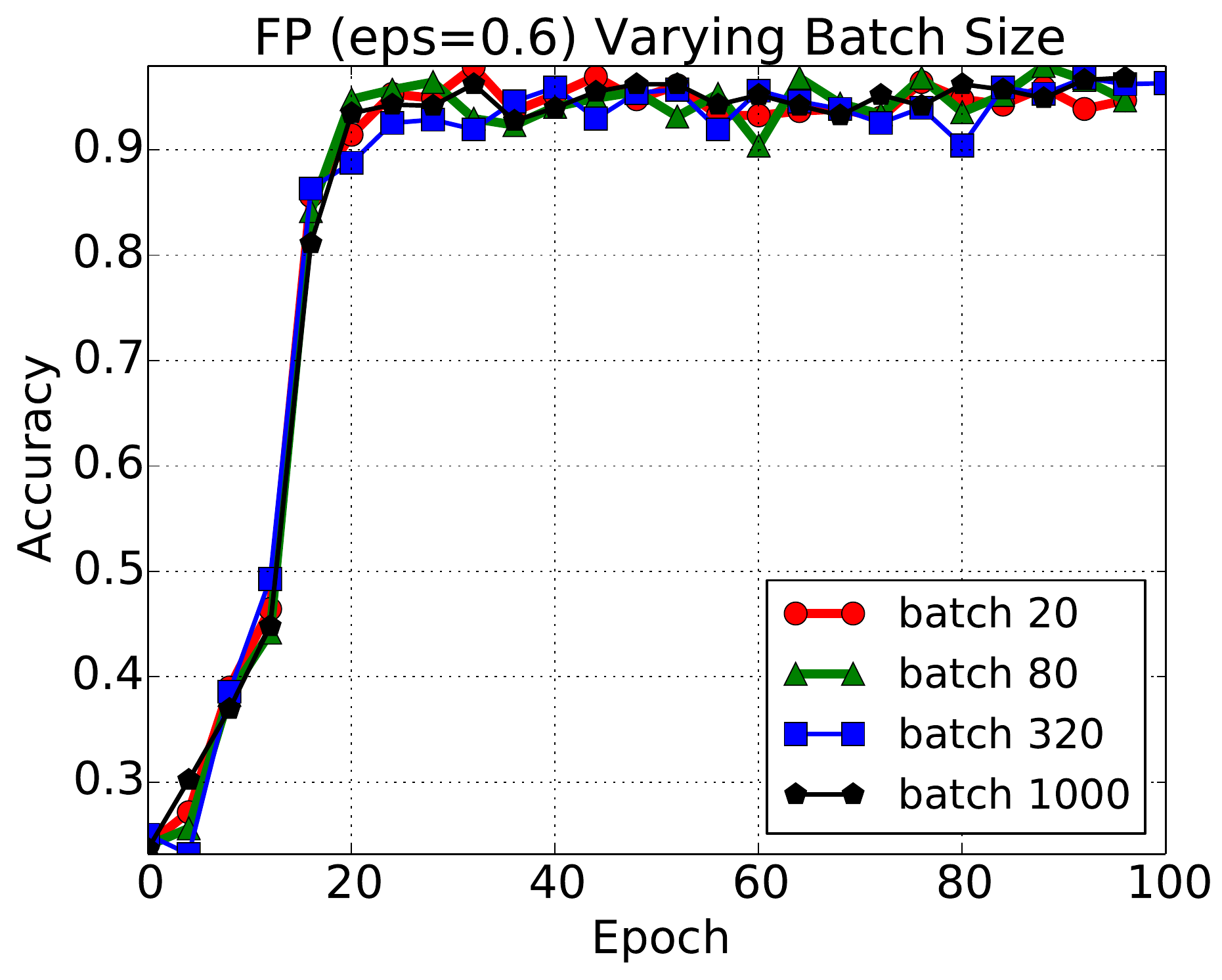}\\
\caption{{\bf  Training epoch vs. test accuracy for bAbI (Task 3) varying exploration $\epsilon$ and batch size.}
Random exploration is important for both reward-based (RBI) and forward prediction (FP).
\label{fig:online-babi-task3}
}
\end{figure*}
\begin{figure*}[h!]
\includegraphics[width=2.5in]{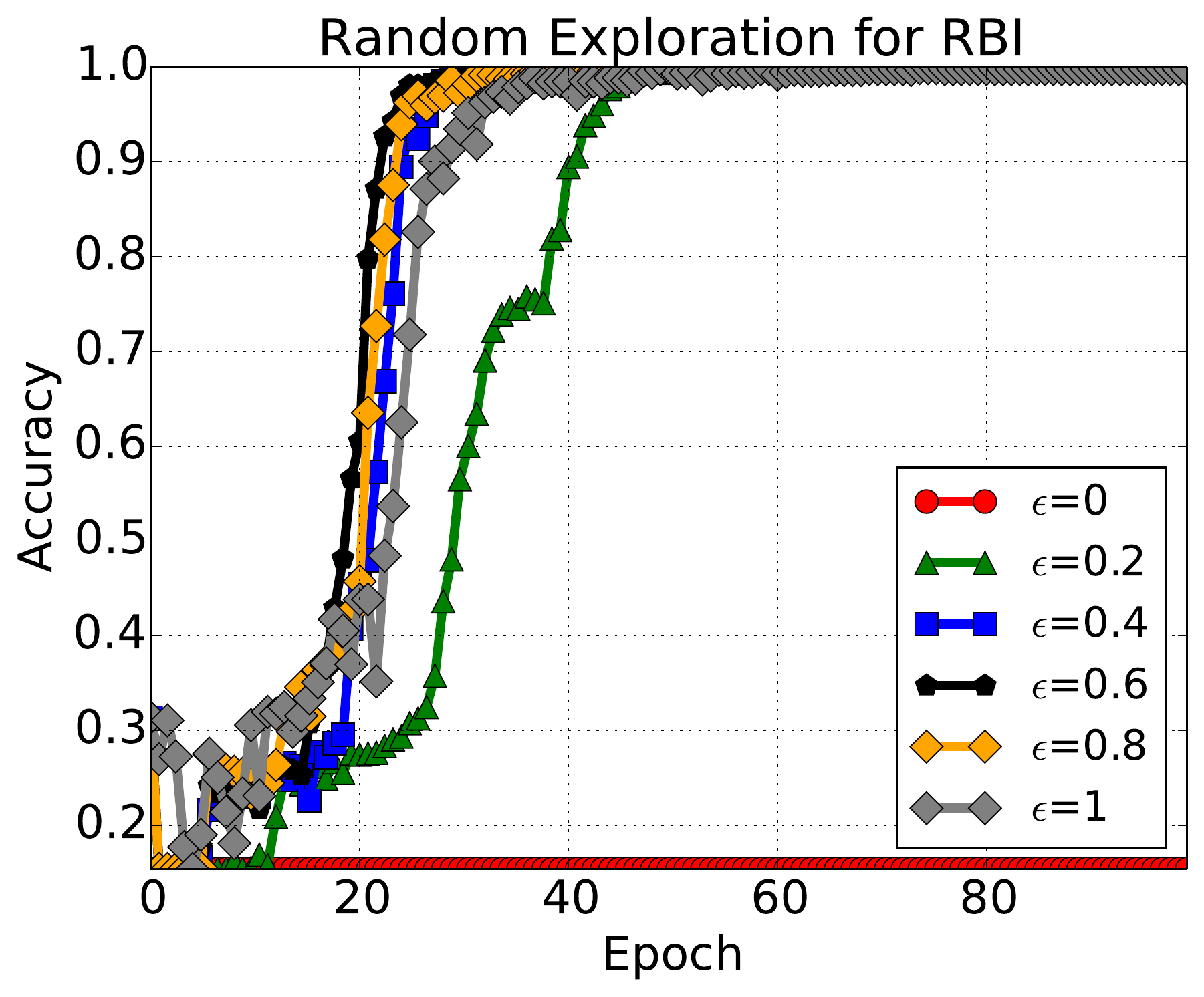}
\includegraphics[width=2.5in]{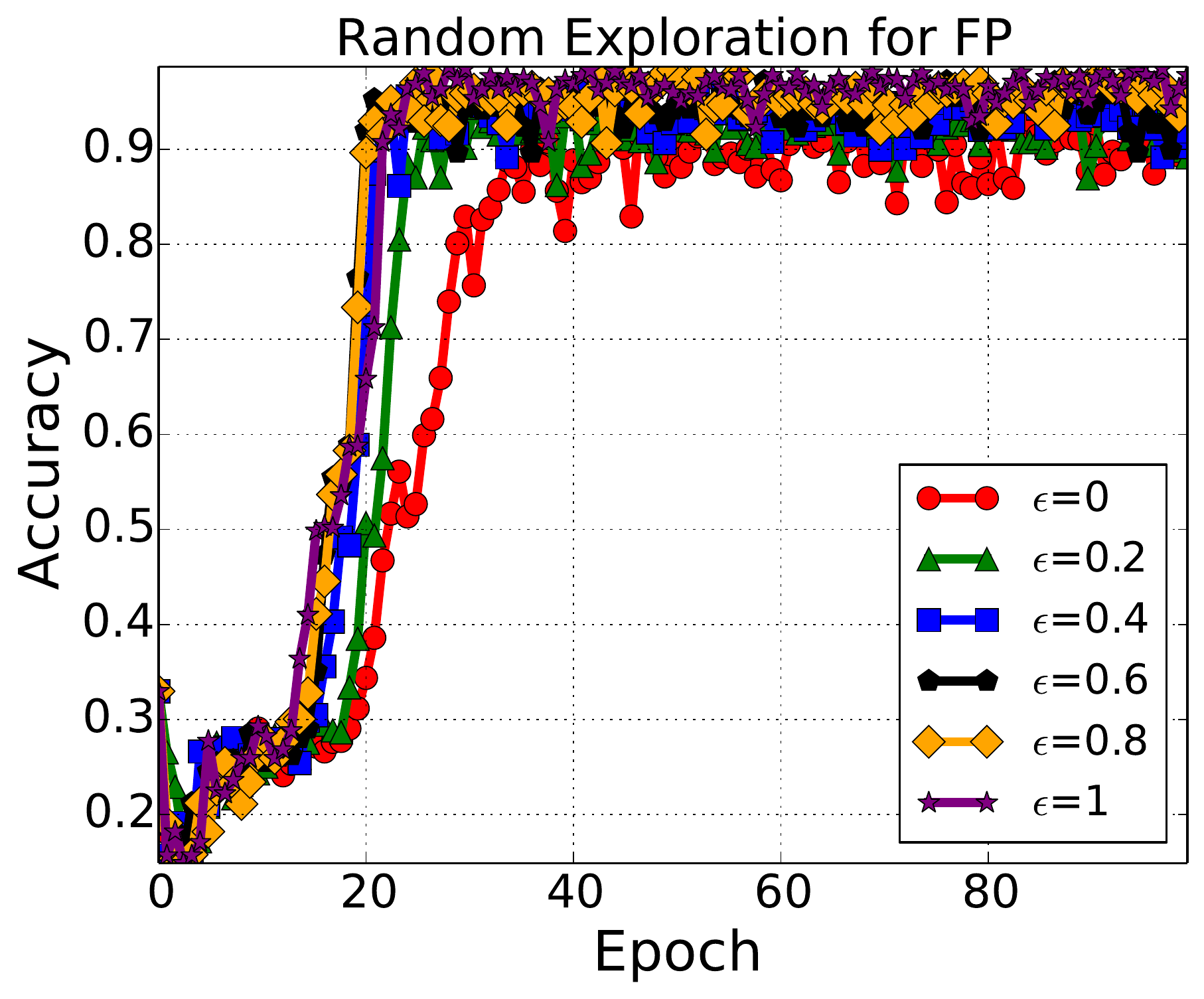}\\
\includegraphics[width=2.5in]{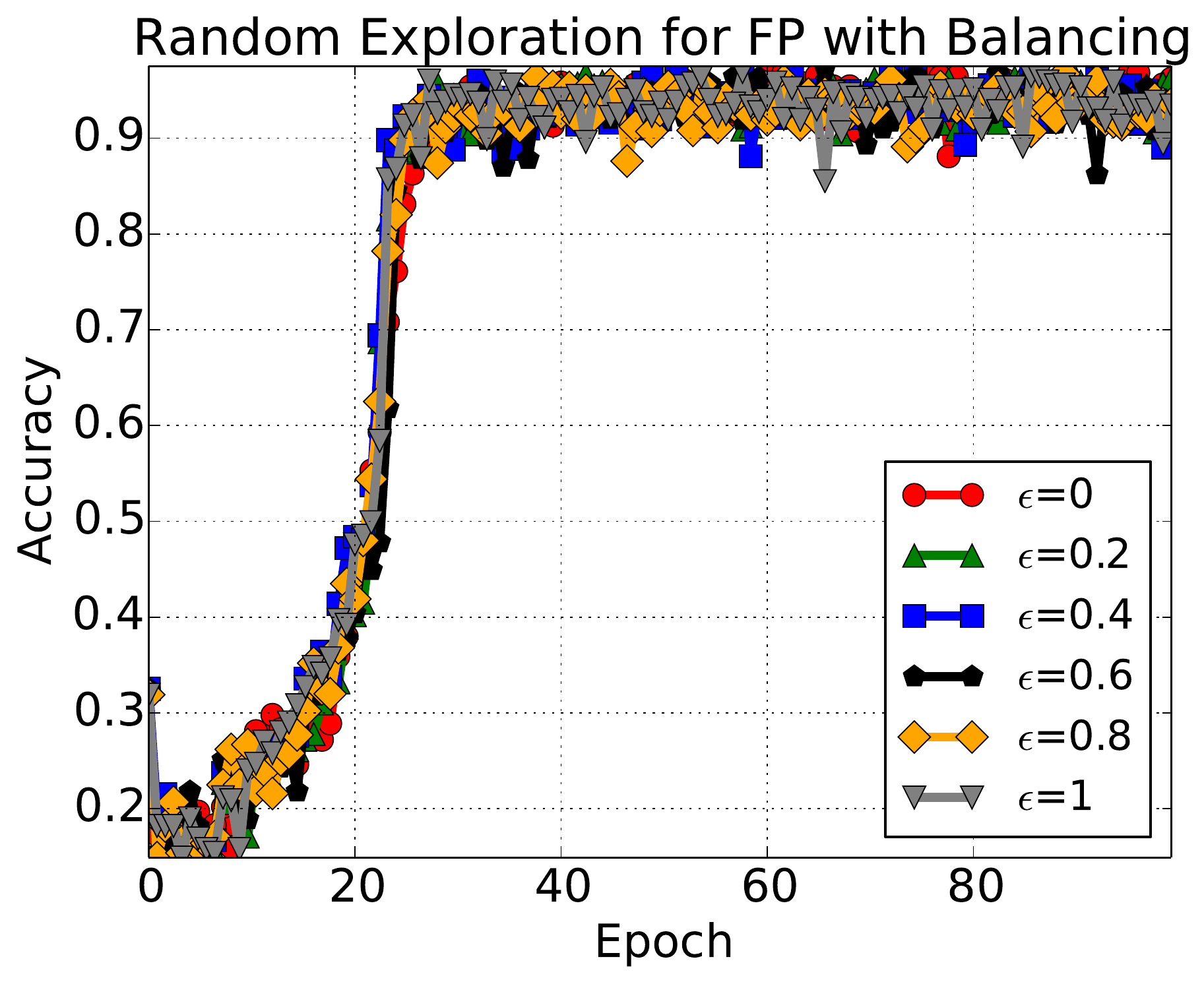}
\includegraphics[width=2.5in]{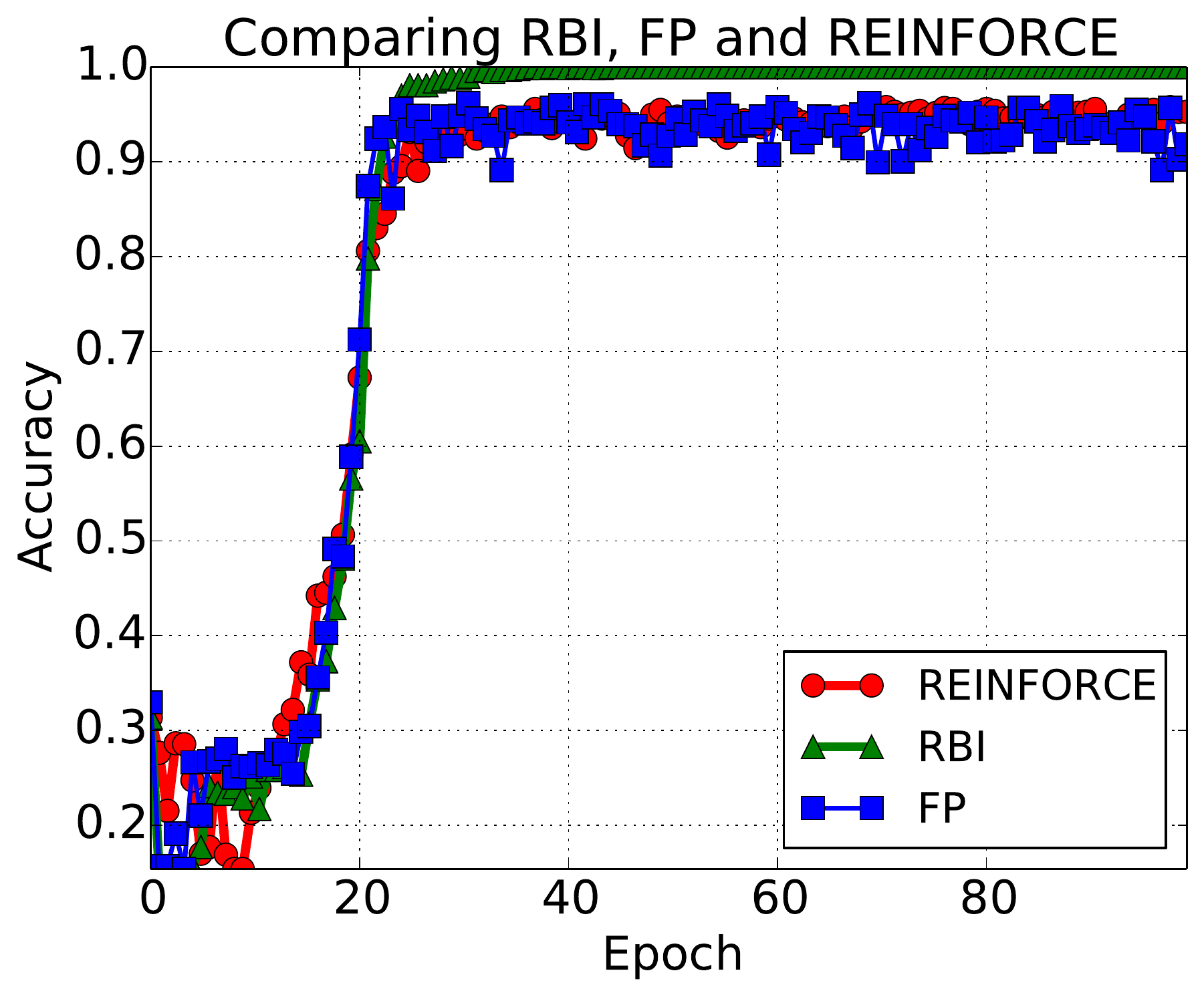}\\
\includegraphics[width=2.5in]{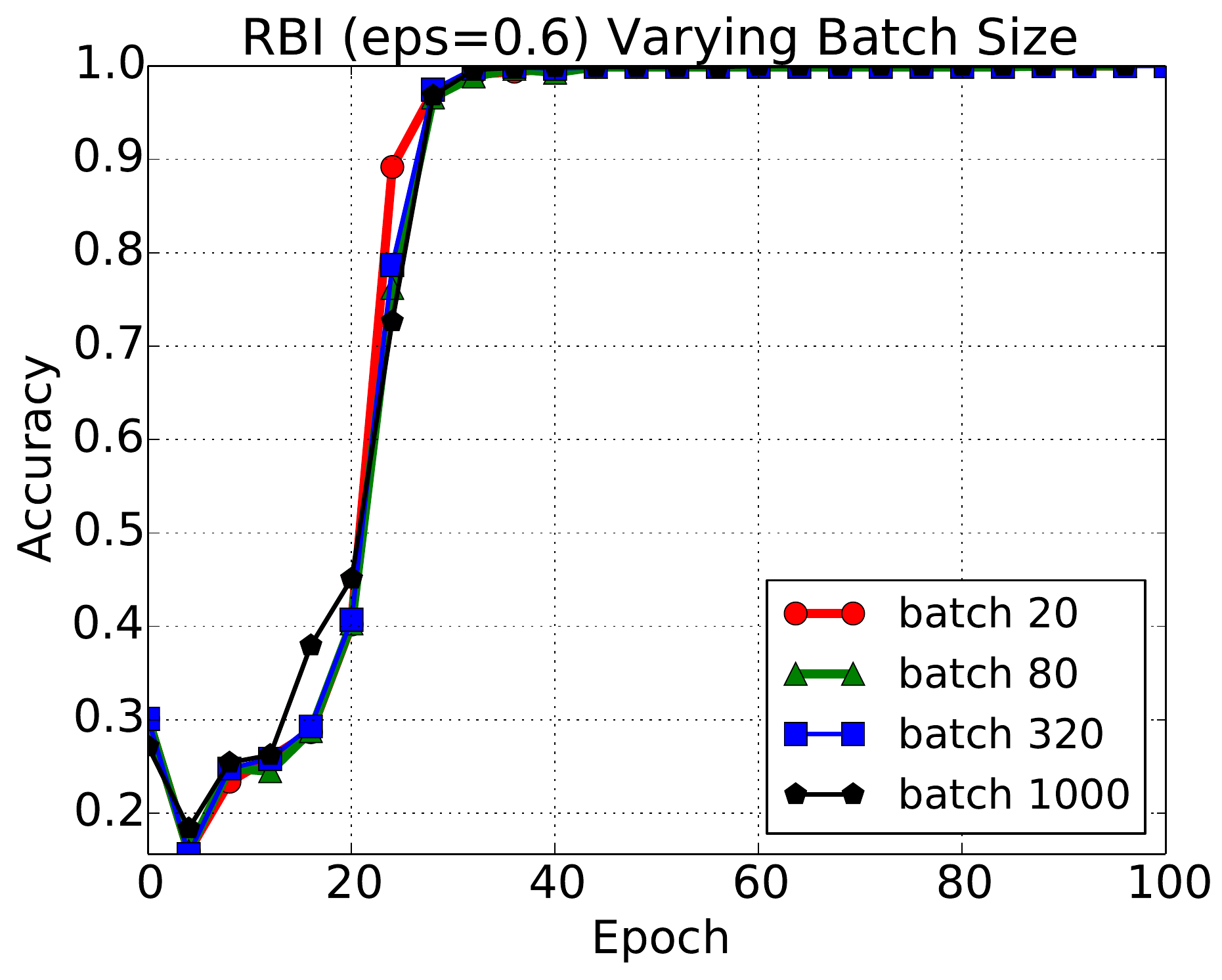}
\includegraphics[width=2.5in]{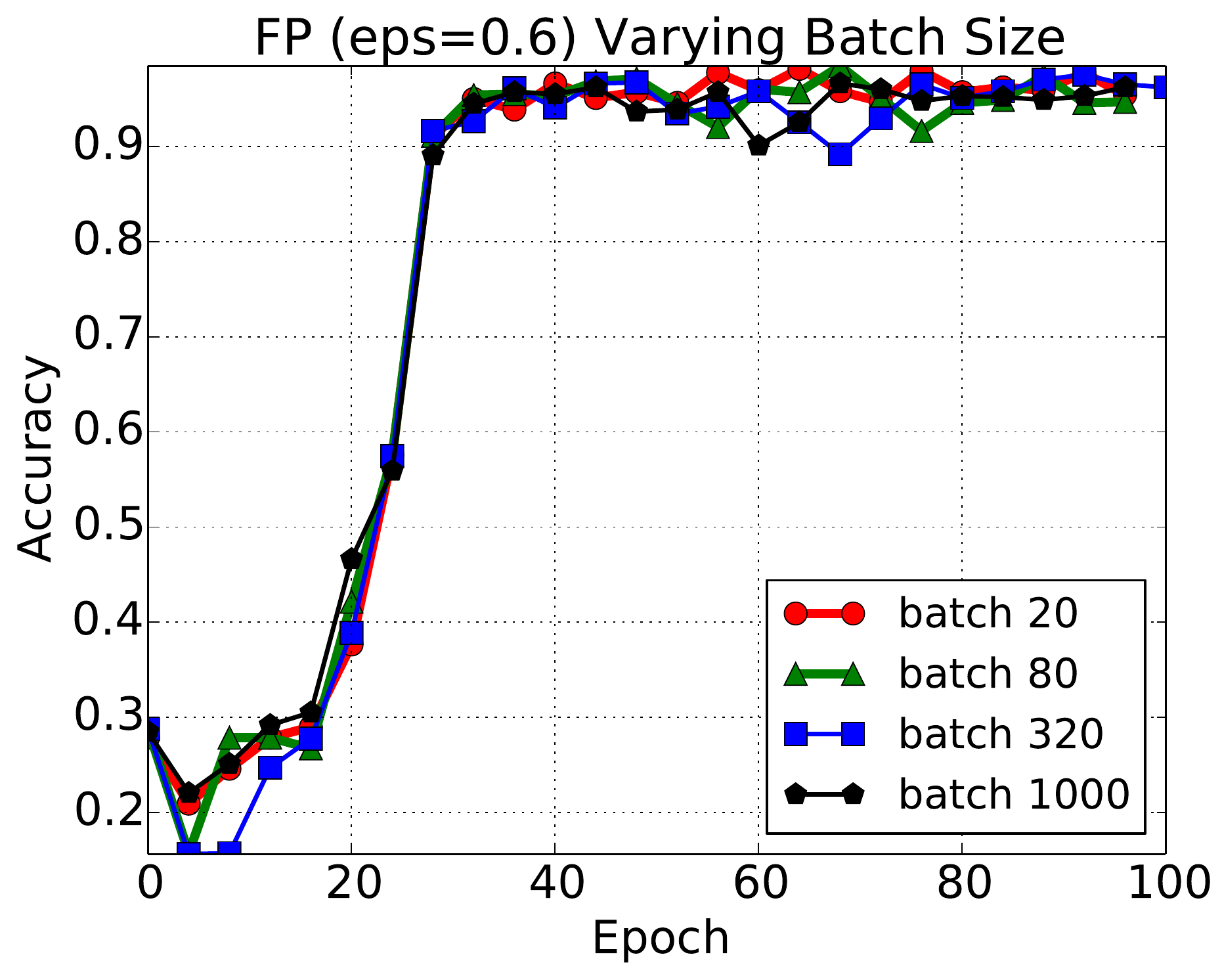}\\
\caption{{\bf  Training epoch vs. test accuracy for bAbI (Task 4) varying exploration $\epsilon$ and batch size.}
Random exploration is important for both reward-based (RBI) and forward prediction (FP).
\label{fig:online-babi-task4}
}
\end{figure*}
\begin{figure*}[h!]
\center
\includegraphics[width=2.35in]{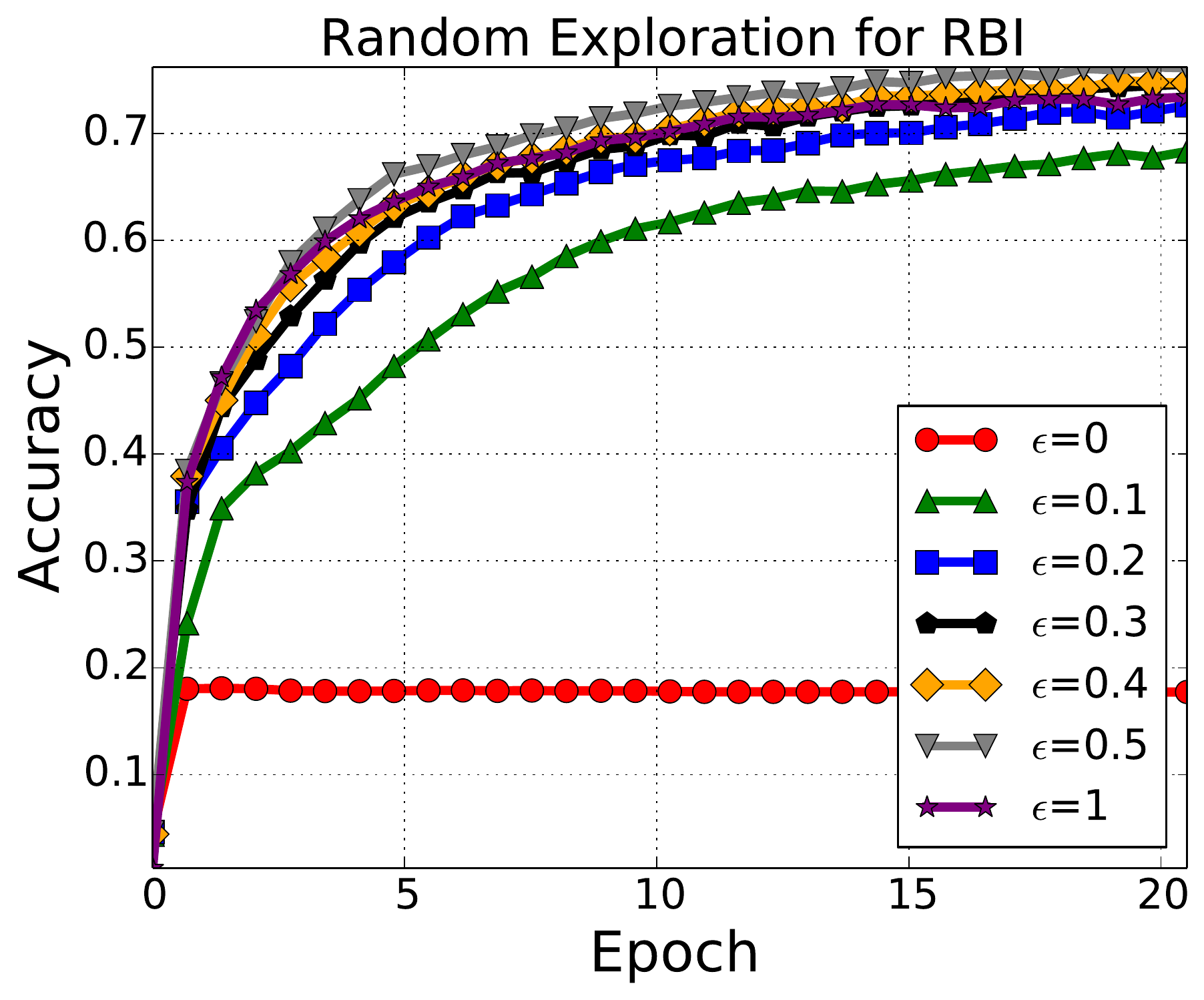}
\includegraphics[width=2.35in]{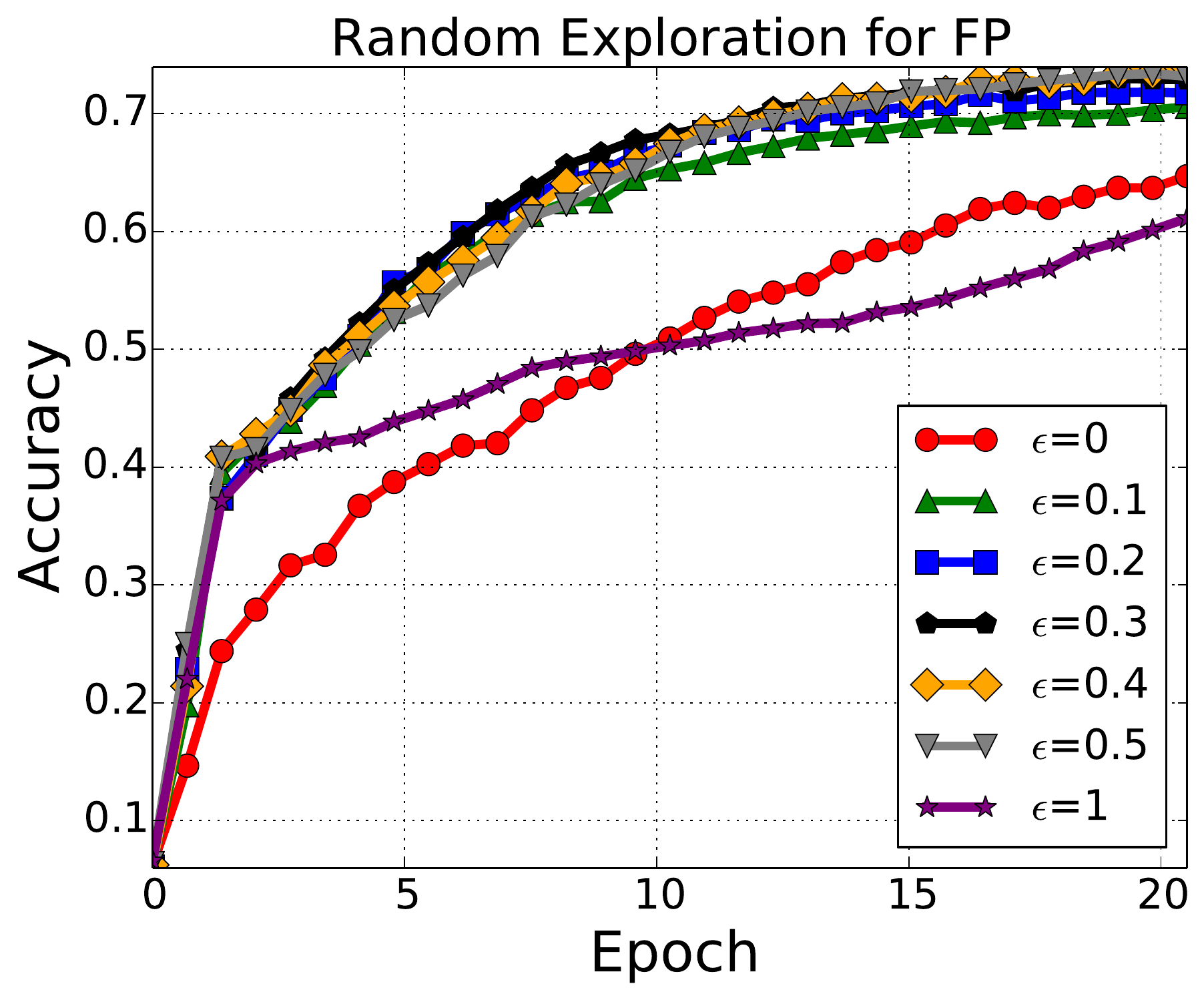}\\
\includegraphics[width=2.35in]{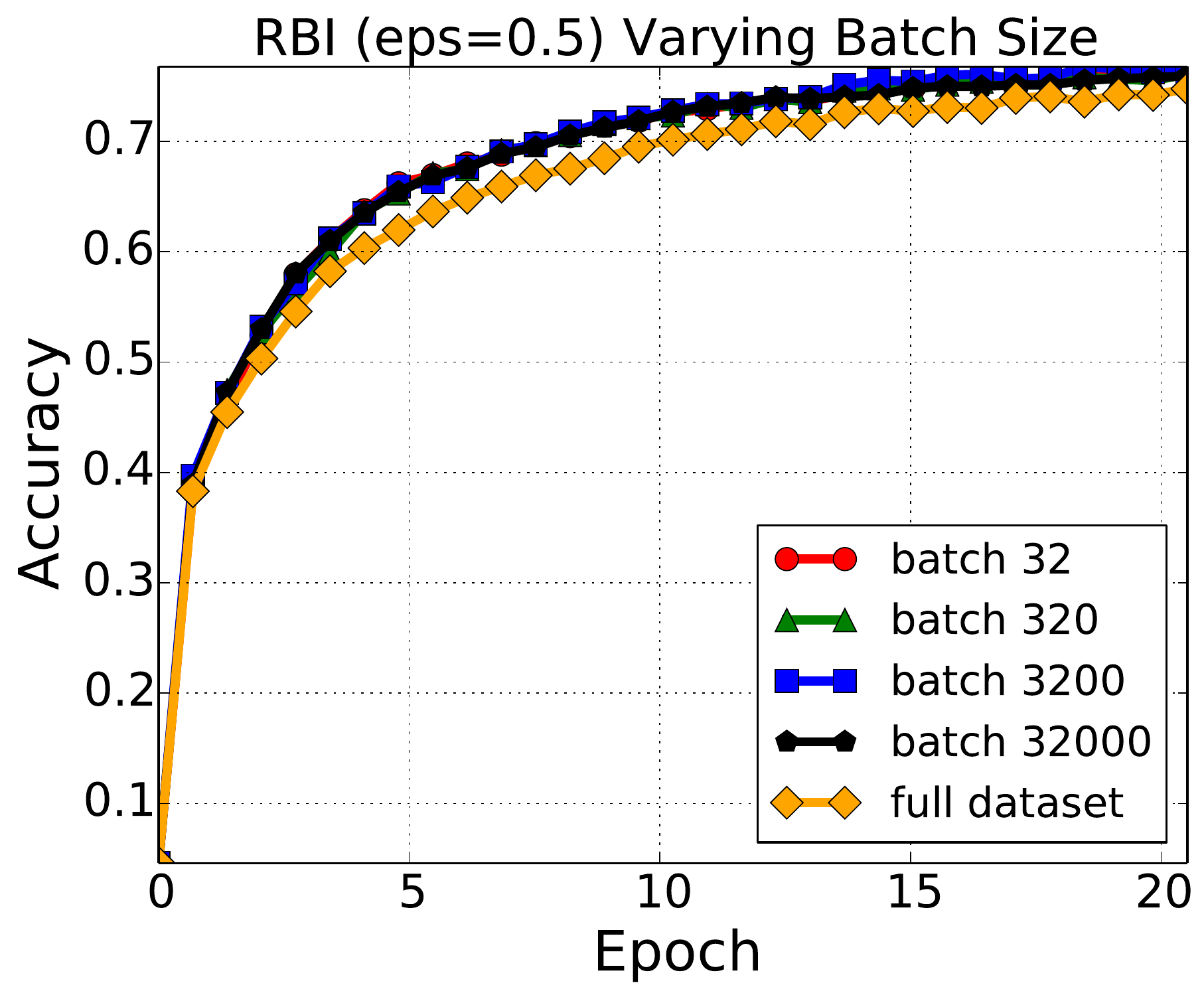}
\includegraphics[width=2.35in]{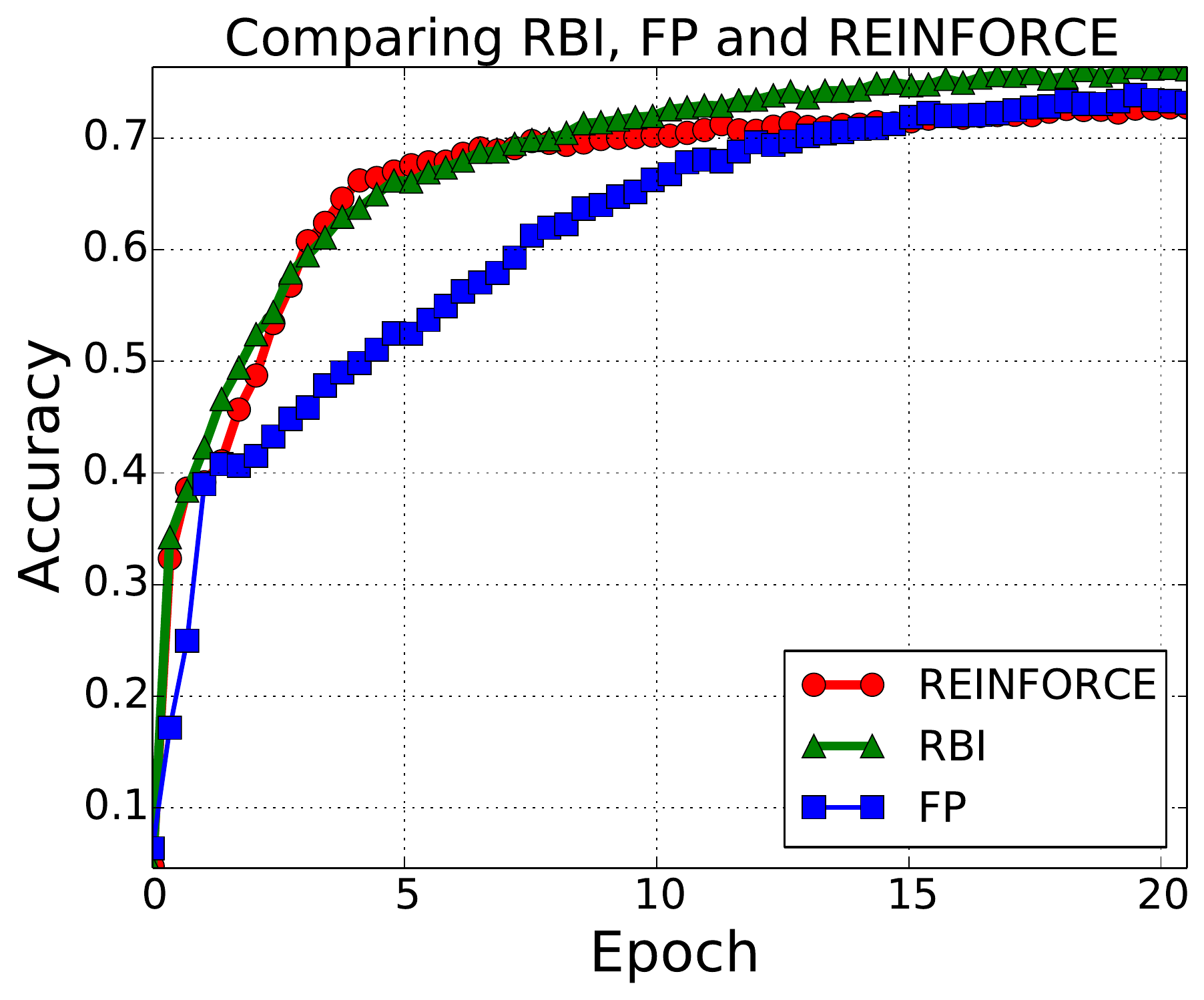}\\
\caption{{\bf WikiMovies}: Training epoch vs. test accuracy on Task $2$ varying (top left panel) exploration rate $\epsilon$ while setting batch size to $32$ for RBI,
(top right panel) for FP,
(bottom left) batch size for RBI, and (bottom right) comparing RBI, REINFORCE and FP
  setting $\epsilon=0.5$. 
The model is robust to the choice of batch size. RBI and REINFORCE perform comparably.
\label{fig:online-movieqa-task2}
}
\end{figure*}
\begin{figure*}[h!]
\center
\includegraphics[width=2.35in]{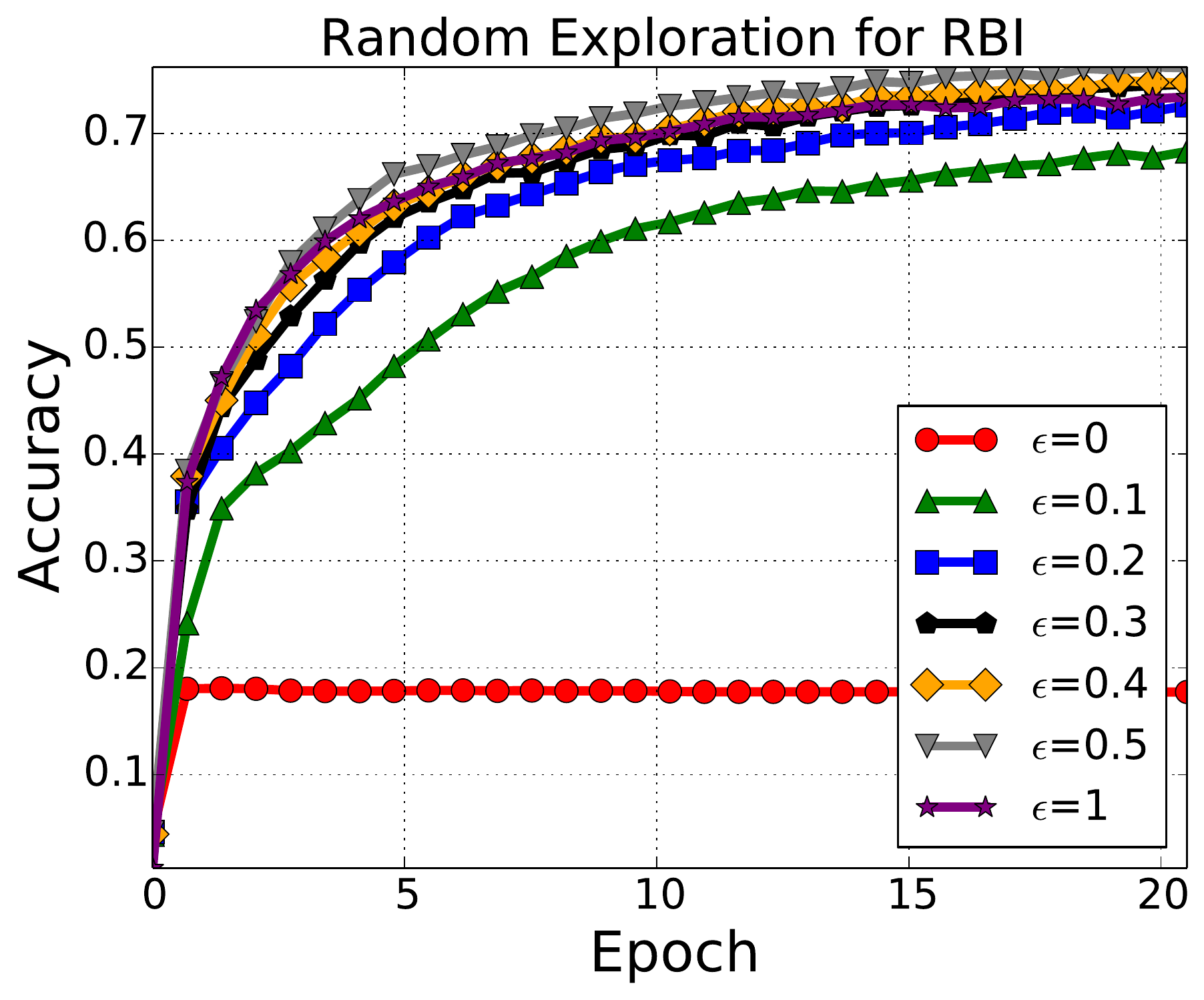}
\includegraphics[width=2.35in]{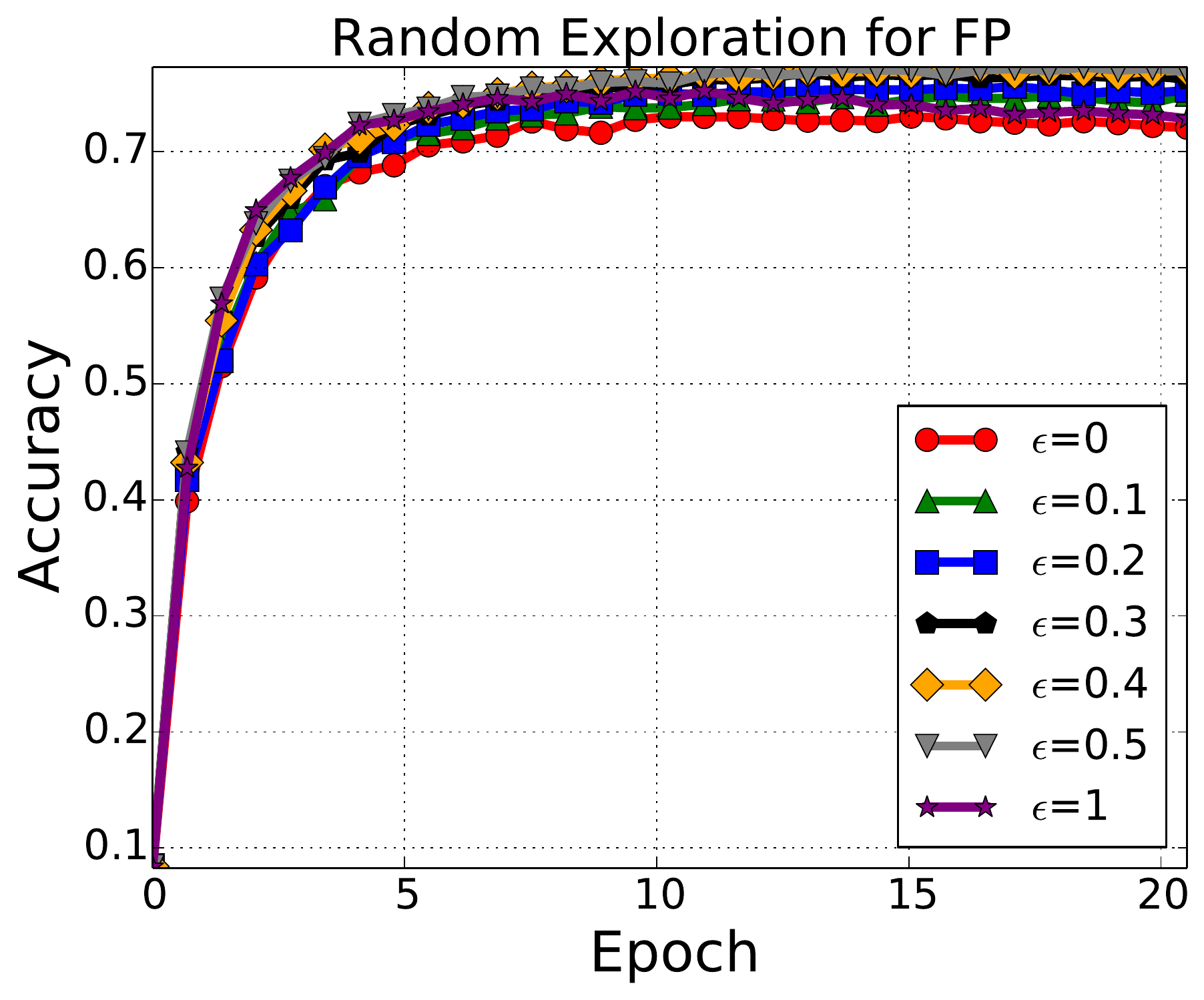}\\
\includegraphics[width=2.35in]{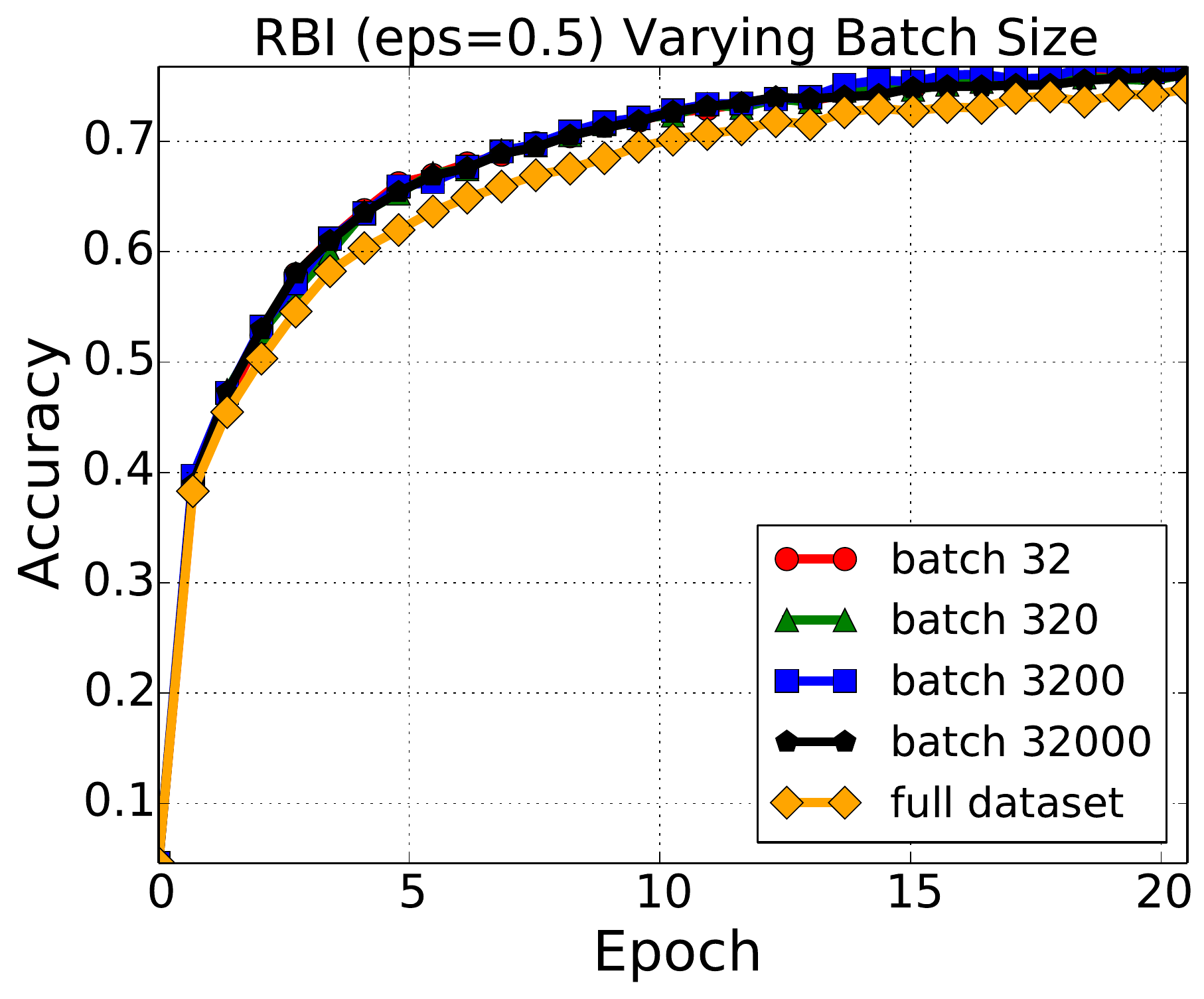}
\includegraphics[width=2.35in]{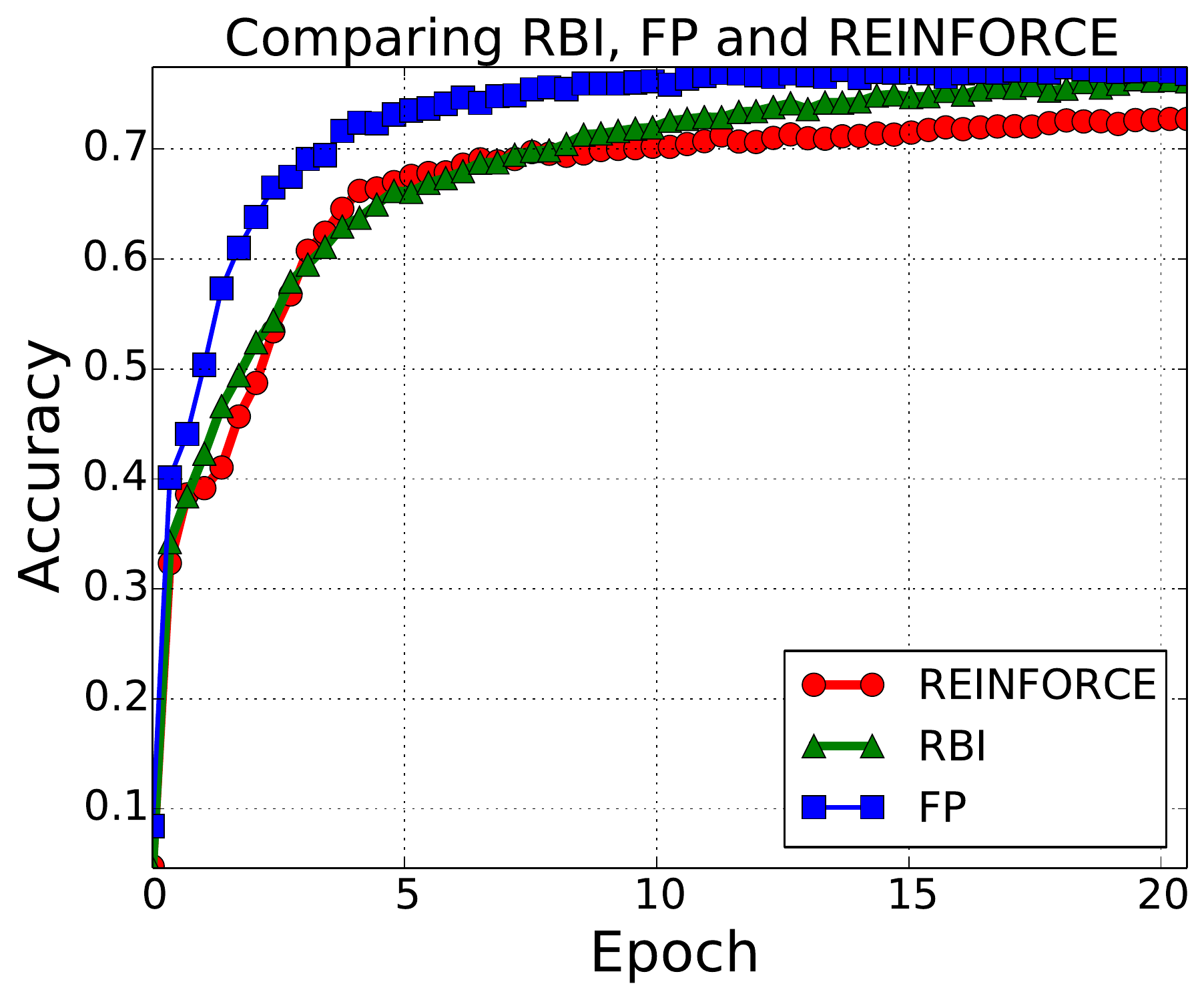}\\
\caption{{\bf WikiMovies}: Training epoch vs. test accuracy on Task $3$ varying (top left panel) exploration rate $\epsilon$ while setting batch size to $32$ for RBI,
(top right panel) for FP,
(bottom left) batch size for RBI, and (bottom right) comparing RBI, REINFORCE and FP
  setting $\epsilon=0.5$. 
The model is robust to the choice of batch size. RBI and REINFORCE perform comparably.
\label{fig:online-movieqa-task3}
}
\end{figure*}
\begin{figure*}[h!]
\center
\includegraphics[width=2.35in]{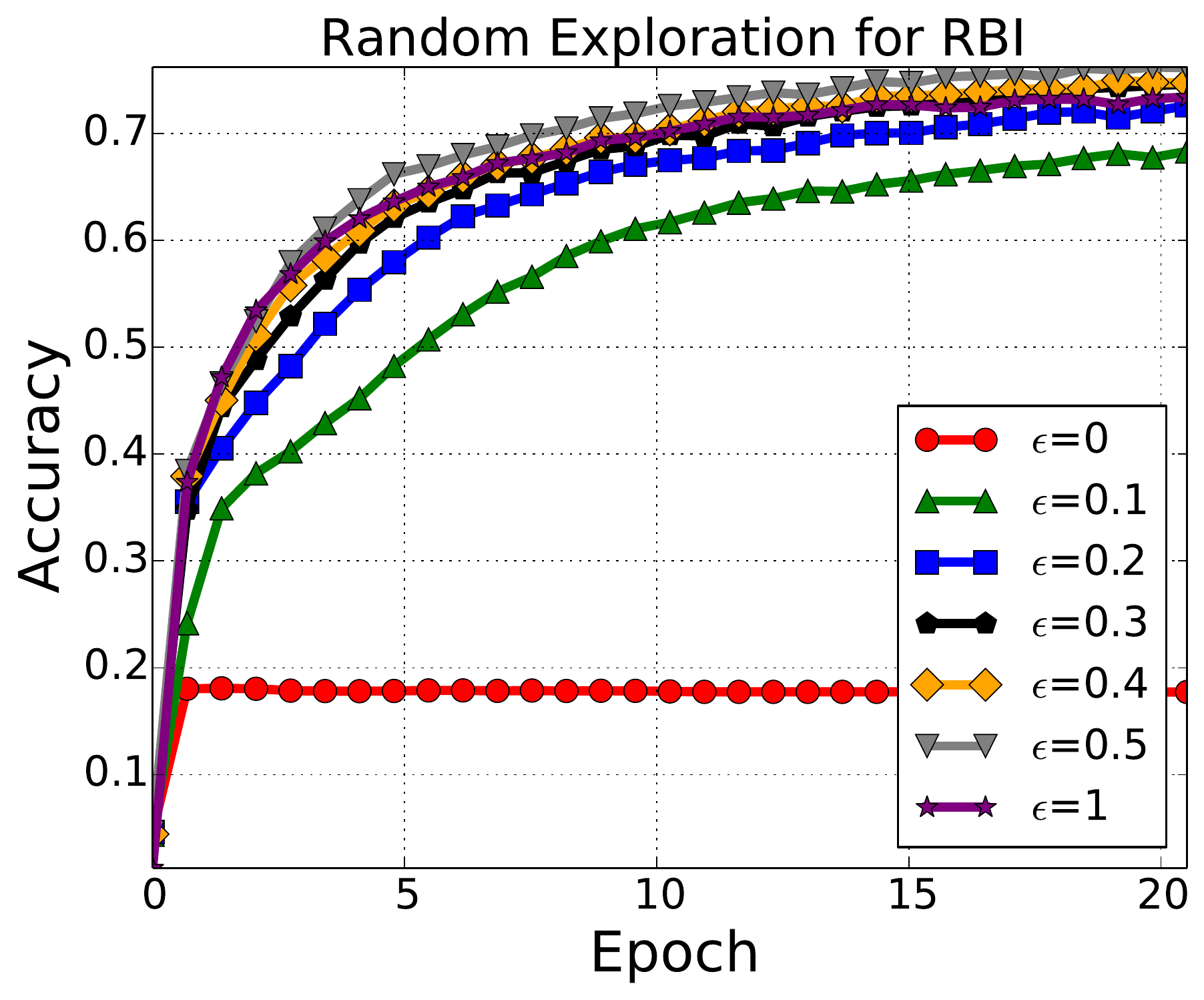}
\includegraphics[width=2.35in]{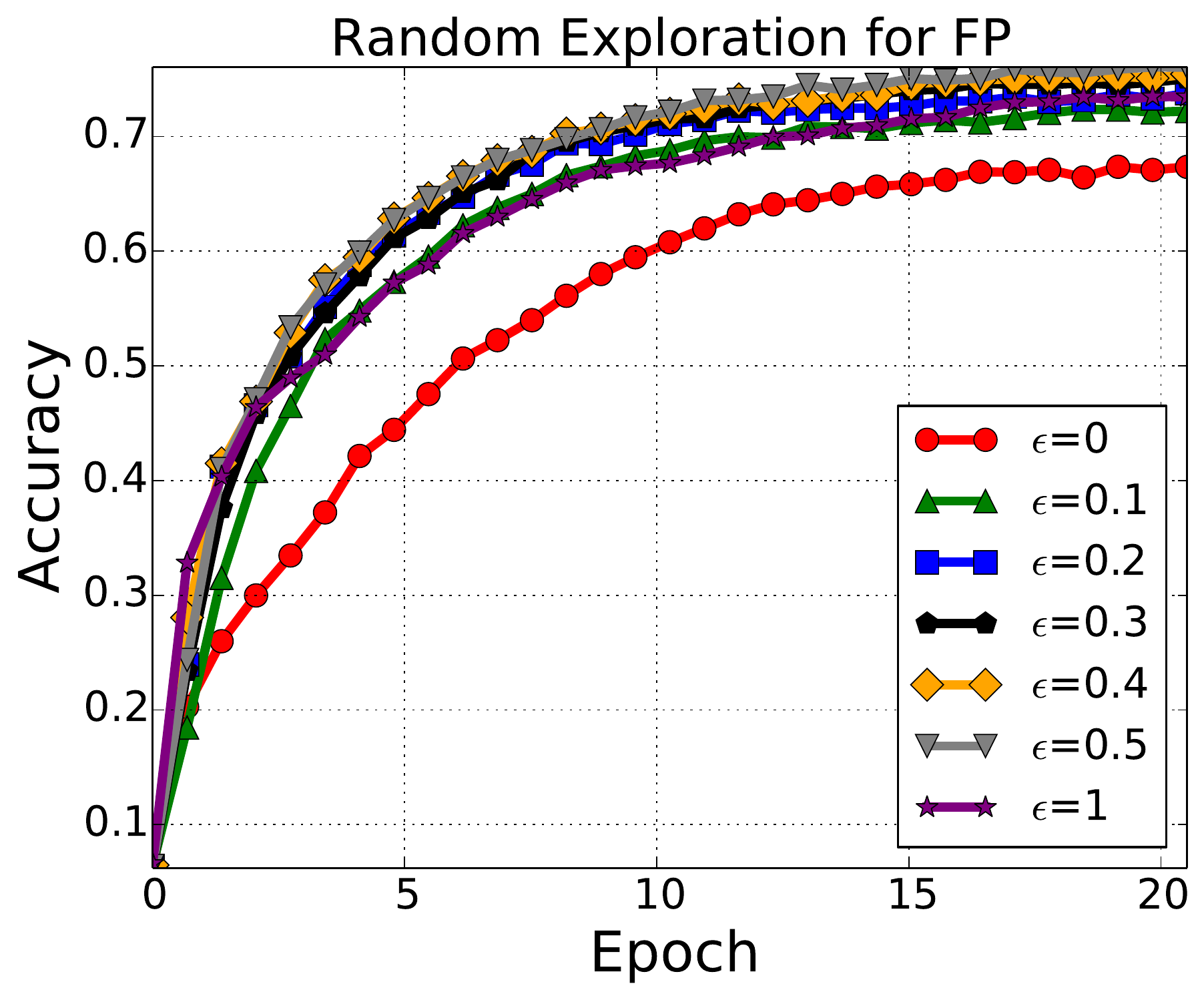}\\
\includegraphics[width=2.35in]{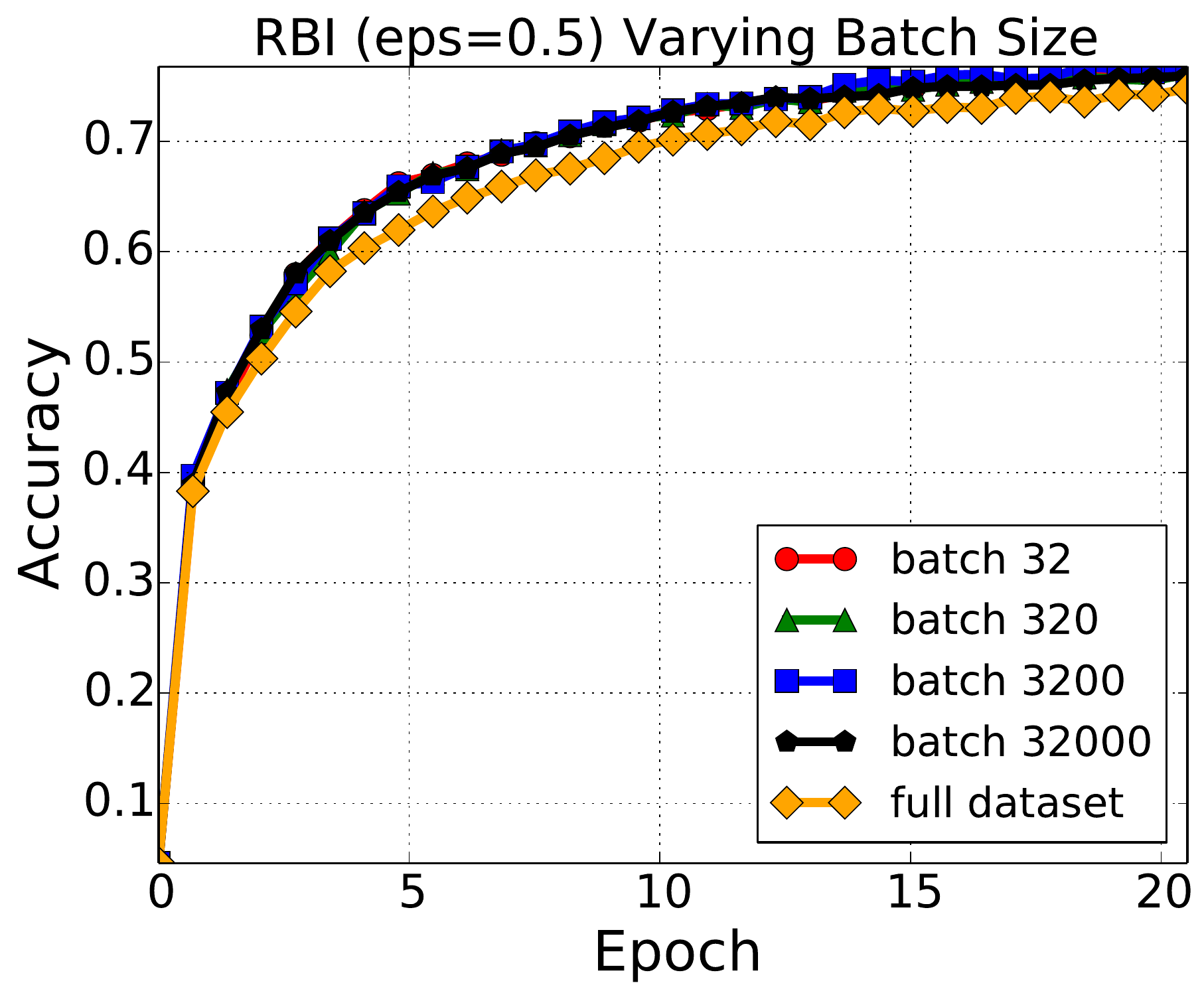}
\includegraphics[width=2.35in]{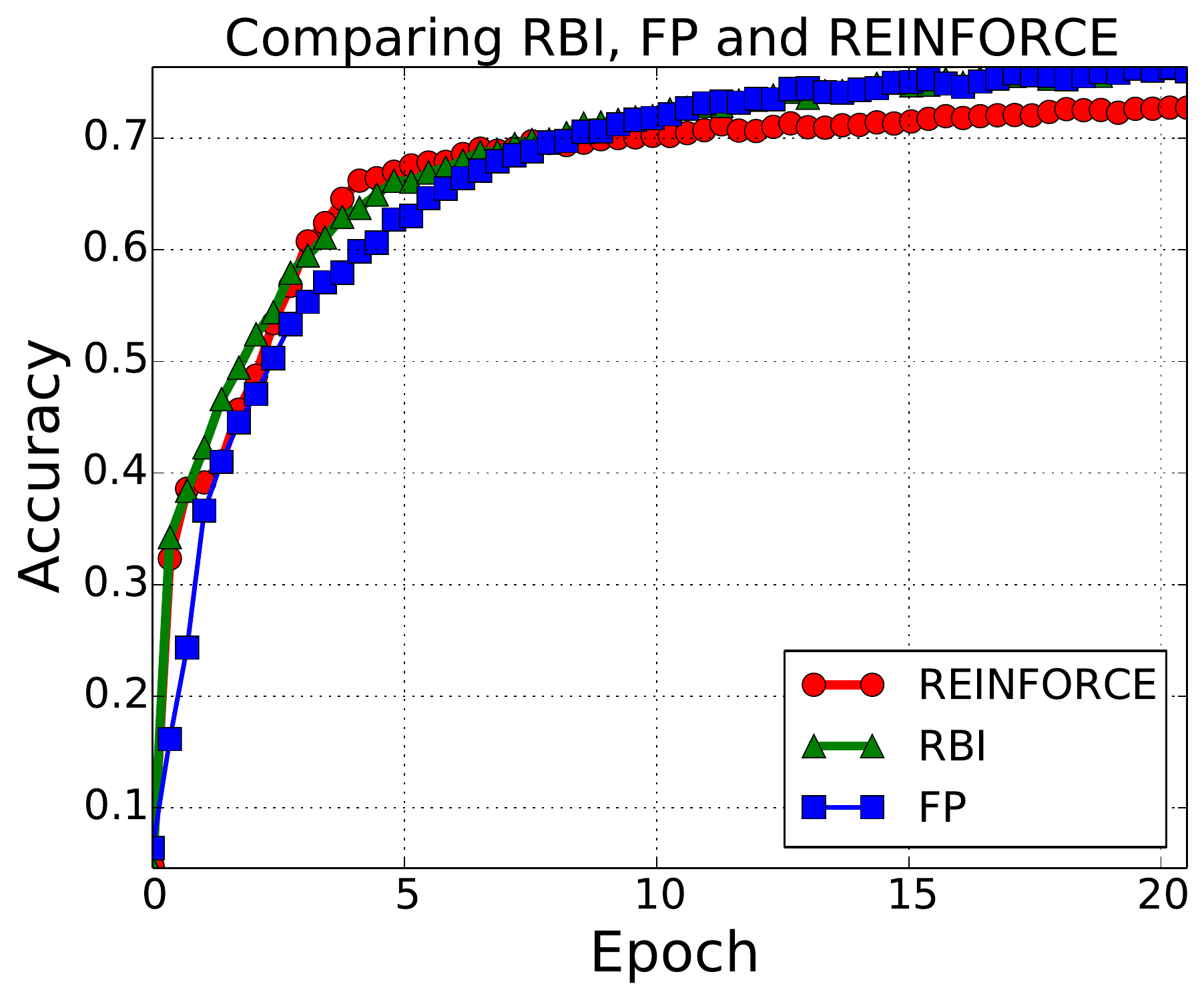}\\
\caption{{\bf WikiMovies}: Training epoch vs. test accuracy on Task $4$ varying (top left panel) exploration rate $\epsilon$ while setting batch size to $32$ for RBI,
(top right panel) for FP,
(bottom left) batch size for RBI, and (bottom right) comparing RBI, REINFORCE and FP
  setting $\epsilon=0.5$. 
The model is robust to the choice of batch size. RBI and REINFORCE perform comparably.
\label{fig:wikitask4}
}
\end{figure*}
\begin{figure*}[h!]
\center
\includegraphics[width=2.35in]{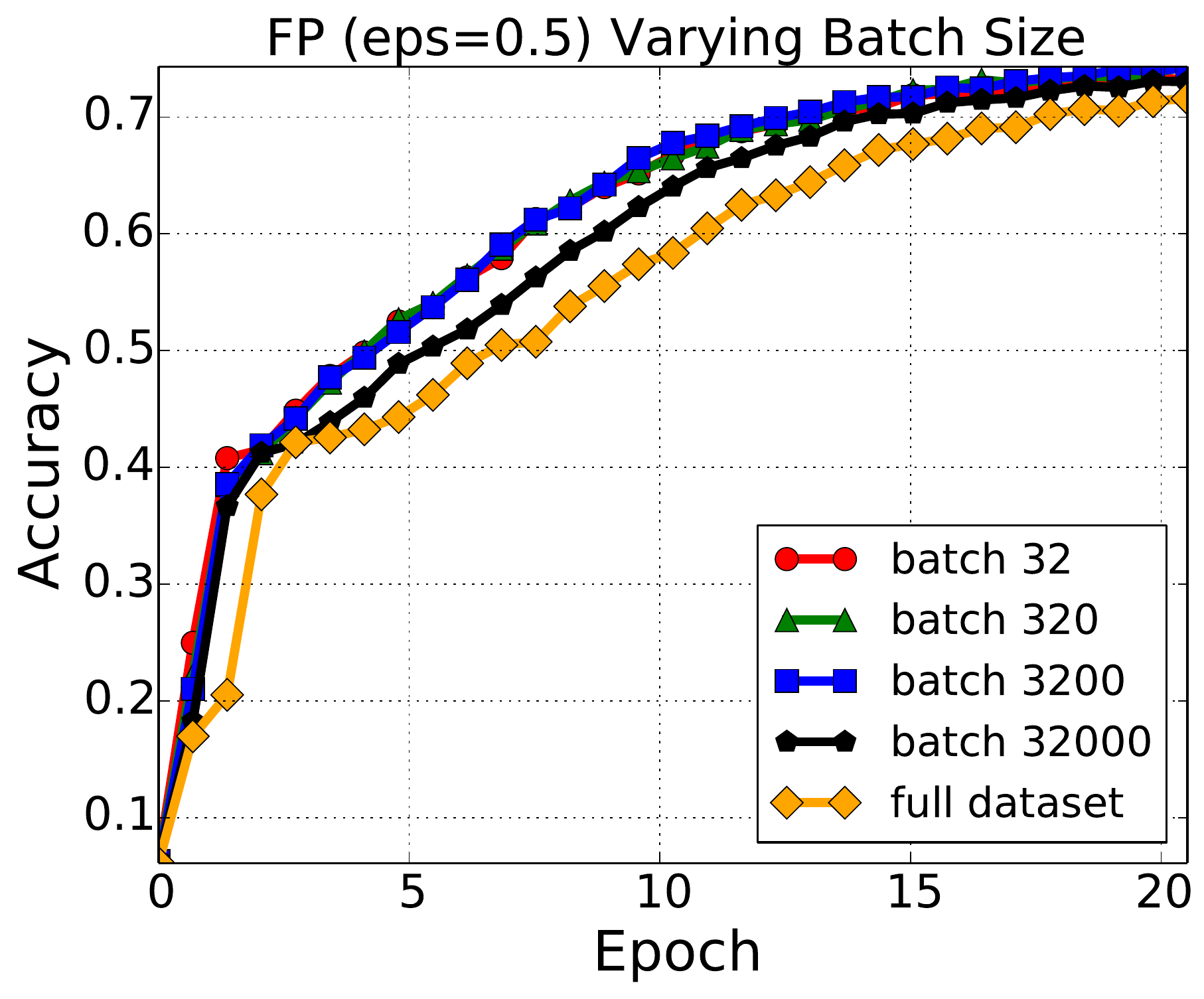}
\includegraphics[width=2.35in]{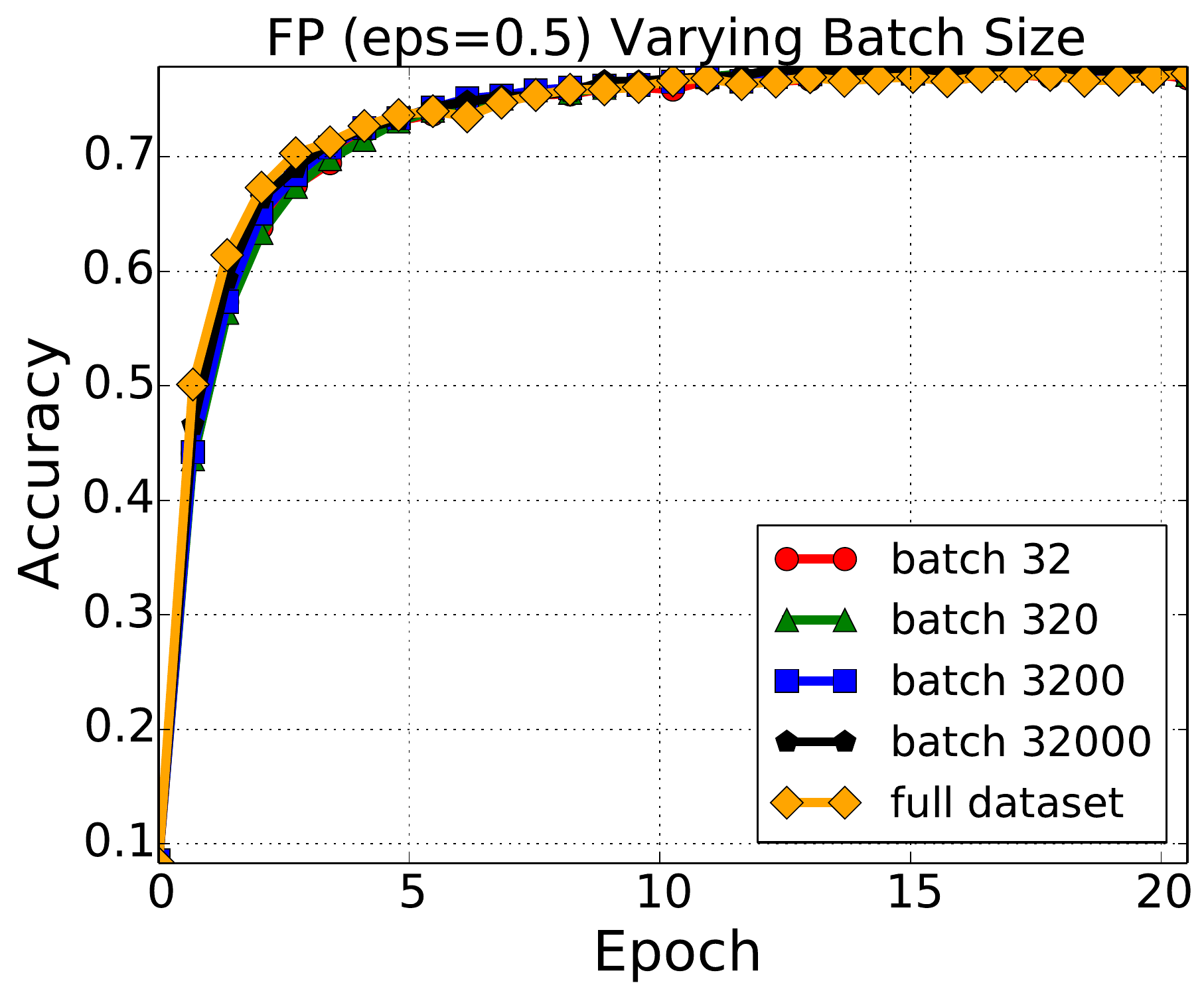}\\
\includegraphics[width=2.35in]{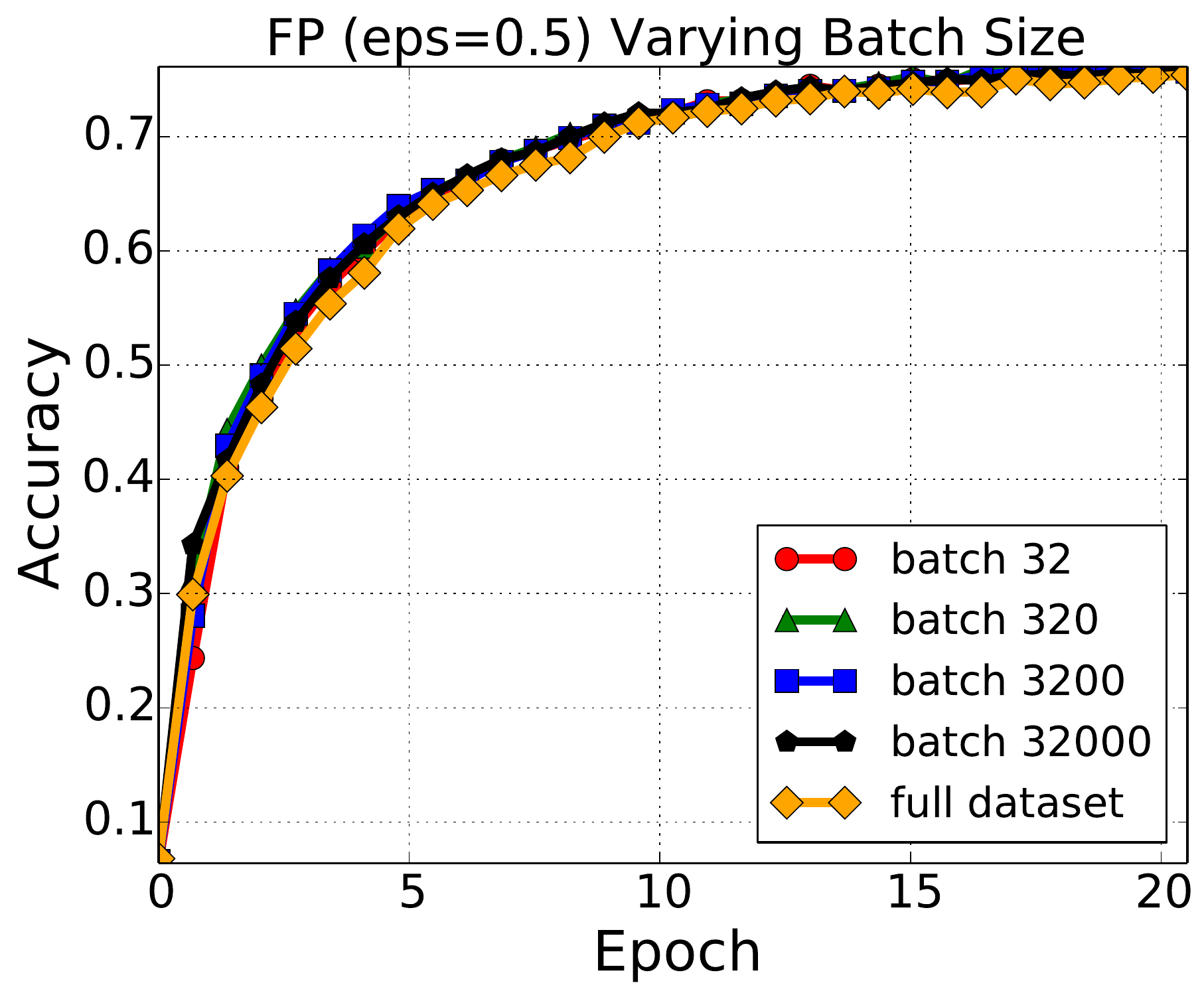}
\includegraphics[width=2.35in]{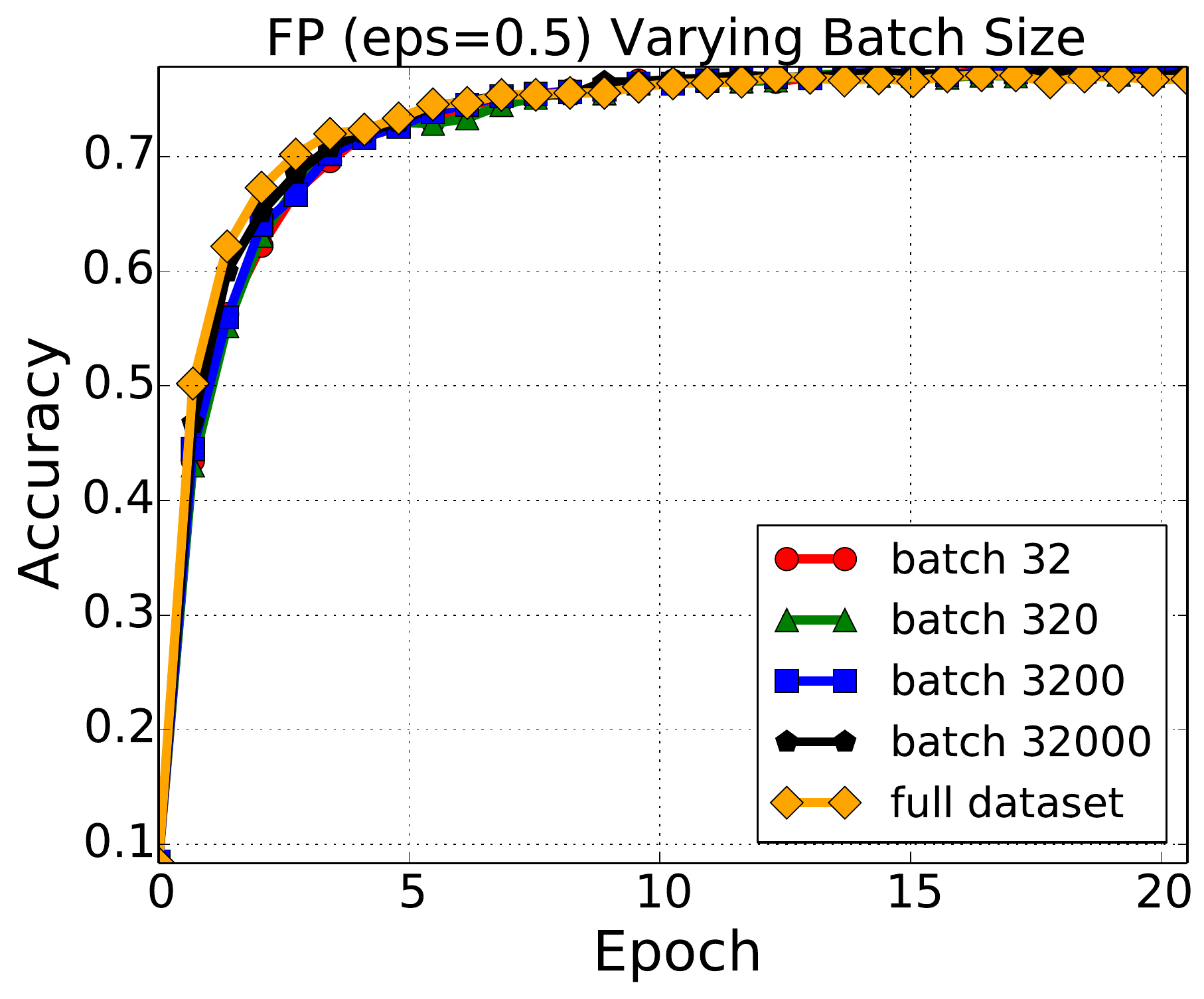}\\
\caption{{\bf WikiMovies}: Training epoch vs. test accuracy with varying  batch size for  FP
on Task 2  (top left panel), 3  (top right panel), 4  (bottom left panel) and 6 (top right panel)
  setting $\epsilon=0.5$. The model is robust to the choice of batch size.
\label{fig:wikitask2}
}
\end{figure*}

\FloatBarrier
\subsection{Additional Experiments For Mechanical Turk Setup}\label{sec:appendix-mturk}

In the experiment in Section 5.2 
 we conducted experiments
with real human feedback. Here, we compare this to a form of synthetic feedback,
mostly as a sanity check, but also to see how much improvement we can get if the
signal is simpler and cleaner (as it is synthetic).
We hence constructed synthetic feedback for the 10,000 responses,
using either Task 2  (positive or negative feedback), Task 3 (answers provided by teacher)
 or a mix (Task 2+3) where we use one or the other for each example
(50\% chance of each).
The latter makes the synthetic data 
have a mixed setup of responses, which more closely mimics the real data case.
The results are given in Table \ref{table:mturk-res-synth}.
The RBI+FP combination is better using the synthetic data than the real data
with Task 2+3 or Task 3, which is to be expected, but the real data is competitive,
despite the difficulty of dealing with its lexical and semantic variability.
The real data is better than using Task 2 synthetic data.

For comparison purposes, we also ran a supervised (imitation learning) MemN2N
on different sized training sets of turker authored questions with gold annotated
labels (so, there are no numerical rewards or textual feedback, this is a pure
 supervised setting).
The results are given in Table \ref{table:mturk-supervised}.
They indicate that RBI+FP and even FP alone get close to the performance of
fully supervised learning.

\begin{table*}[!tbh]
\begin{center}
\begin{tabular}{l|c|c|c|c|}
Model                         & $r=0$   &  $r=0.1$  &  $r=0.5$  & $r=1$ \\
\hline
Reward Based Imitation (RBI)         & 0.333    &  0.340   &  0.365   & 0.375 \\
Forward Prediction (FP) [real]       & 0.358    &  0.358   &  0.358   & 0.358 \\
RBI+FP                  [real]       & 0.431    &  0.438   &  0.443   & 0.441 \\
\hline
Forward Prediction (FP) [synthetic Task 2]  & 0.188    &  0.188   &  0.188   & 0.188 \\
Forward Prediction (FP) [synthetic Task 2+3]  & 0.328    &  0.328   &  0.328   & 0.328 \\
Forward Prediction (FP) [synthetic Task 3]  & 0.361    &  0.361   &  0.361   & 0.361 \\
\hline
RBI+FP                  [synthetic Task 2]  & 0.382    &  0.383   &  0.407   & 0.408 \\
RBI+FP                  [synthetic Task 2+3]& 0.459    &  0.465   &  0.464   & 0.478 \\
RBI+FP                  [synthetic Task 3]  & 0.473    &  0.486   &  0.490   &  0.494\\
\end{tabular}
\end{center}
\caption{{\bf Incorporating Feedback From Humans via Mechanical Turk: comparing real human feedback to synthetic feedback.}
Textual feedback is provided for 10,000 model predictions (from a model trained with 1k labeled training examples), and additional sparse
binary rewards (fraction $r$ of examples have rewards).
We compare real feedback (rows 2 and 3) to synthetic feedback when using FP or RBI+FP (rows 4 and 5).
\label{table:mturk-res-synth}
}
\end{table*}

\begin{table*}[!tbh]
\begin{center}
\begin{tabular}{l|c|c|c|c|c|}
Train data size     & 1k    &  5k &  10k & 20k & 60k \\
\hline
Supervised MemN2N   & 0.333 & 0.429 & 0.476 & 0.526 & 0.599 \\
\end{tabular}
\end{center}
\caption{{\bf Fully Supervised (Imitation Learning) Results on Human Questions}
\label{table:mturk-supervised}
}
\end{table*}

\begin{table*}[!h]
\begin{center}
\begin{tabular}{l|c|c|c|c|c|}
           & $r=0$   &  $r=0.1$  &  $r=0.5$  & $r=1$ \\   
\hline
$\epsilon=0 $   & 0.499   & 0.502     & 0.501    & 0.502 \\
$\epsilon=0.1$  & 0.494   & 0.496     & 0.501    & 0.502 \\
$\epsilon=0.25$ & 0.493   & 0.495     & 0.496    & 0.499 \\ 
$\epsilon=0.5$  & 0.501   & 0.499     & 0.501    & 0.504 \\
$\epsilon=1$    & 0.497   & 0.497     & 0.498    & 0.497 \\
\end{tabular}
\end{center}
\caption{{\bf Second Iteration of Feedback}
Using synthetic textual feedback of synthetic Task2+3 with the RBI+FP method,
an additional iteration of data collection of 10k examples,
varying sparse binary reward fraction $r$ and exploration $\epsilon$.
The performance of the first iteration model was 0.478.
\label{table:mturk-2nd-it}
}
\end{table*}

\subsection{Second Iteration of Feedback}

We conducted experiments with an additional iteration of data collection
for the case of binary rewards and textual feedback using the synthetic Task 2+3 mix.
We selected the best model from the previous training, using RBI+FP with $r=1$
which previously gave a test accuracy of 0.478 (see Table \ref{table:mturk-res-synth}).
Using that model as a predictor, we collected an additional 10,000 training 
examples.
We then continue to train our model using the original 1k+10k training set,
plus the additional 10k. As before, we report the test accuracy
varying $r$ on the additional collected set. We also report the performance
from varying $\epsilon$, the proportion of random exploration of predictions on the
new set. The results are reported in Table \ref{table:mturk-2nd-it}.
Overall, performance is improved in the second iteration,
with slightly better performance for large $r$ and $\epsilon=0.5$.
However,  the improvement is mostly invariant to those parameters, likely 
because FP takes advantage of feedback from incorrect predictions in any case.